\documentclass[table]{article} %
\usepackage{collas2025_conference,times}
\usepackage{easyReview}

\usepackage{hyperref}
\hypersetup{
    colorlinks=true,
    linkcolor=red,
    filecolor=magenta,
    urlcolor=blue,
    citecolor=purple,
    pdftitle={Overleaf Example},
    pdfpagemode=FullScreen,
    }

\usepackage{mathtools}   %
\usepackage{amssymb}     %
\newcommand\norm[1]{\left\lVert#1\right\rVert}

\usepackage{multirow}
\usepackage{graphicx}
\usepackage{subcaption}  %
\usepackage{comment}     %
\usepackage{booktabs}    %
\usepackage{multirow}    %
\usepackage{tikz}
\usepackage{csquotes}
\usetikzlibrary{positioning}
\usetikzlibrary{decorations.pathreplacing}

\definecolor{C0}{HTML}{1f77b4}
\definecolor{C1}{HTML}{ff7f0e}
\definecolor{C2}{HTML}{2ca02c}
\definecolor{C3}{HTML}{d62728}
\definecolor{C4}{HTML}{9467bd}
\definecolor{C5}{HTML}{8c564b}
\definecolor{C6}{HTML}{e377c2}
\definecolor{C7}{HTML}{7f7f7f}
\definecolor{C8}{HTML}{bcbd22}
\definecolor{C9}{HTML}{17becf}

\definecolor{V0}{HTML}{fde725}
\definecolor{V1}{HTML}{b5de2b}
\definecolor{V2}{HTML}{6ece58}
\definecolor{V3}{HTML}{35b779}
\definecolor{V4}{HTML}{1f9e89}
\definecolor{V5}{HTML}{26828e}
\definecolor{V6}{HTML}{31688e}
\definecolor{V7}{HTML}{3e4989}
\definecolor{V8}{HTML}{482878}
\definecolor{V9}{HTML}{440154}

\definecolor{mygray}{RGB}{230,230,230} %

\usepackage{pifont} %

\newcommand{\solidline}[1]{\textcolor{#1}{\rule[0.5ex]{0.6cm}{1pt}}} %
\newcommand{\dashedline}[1]{\textcolor{#1}{\rule[0.5ex]{0.15cm}{1pt}~\rule[0.5ex]{0.15cm}{1pt}~\rule[0.5ex]{0.15cm}{1pt}\;}} %

\newcommand\vbar{\kern1pt\rule[-0.3ex]{1pt}{2ex}\kern1pt}
\newcommand\vbars{\textcolor{V9}{\vbar}\textcolor{V8}{\vbar}\textcolor{V6}{\vbar}\textcolor{V4}{\vbar}\textcolor{V2}{\vbar}\textcolor{V0}{\vbar}}

\newcommand{\figmark}[1]{\textcolor{gray}{\sffamily \textbf{#1}}}

\usepackage{marvosym} %

\title{How Weight Resampling and Optimizers Shape the Dynamics of Continual Learning and Forgetting in Neural Networks}

\author{Lapo Frati\\
University of Vermont \\
\texttt{lfrati@uvm.edu} \\
\And
Neil Traft  \\
University of Vermont \\
\texttt{ntraft@uvm.edu} \\
\And
Jeff Clune \\
University of British Columbia \\
Vector Institute \\ 
Canada CIFAR AI Chair \\
\And
Nick Cheney \\
University of Vermont \\
\texttt{ncheney@uvm.edu}
}

\preprintcopy %

\begin{document}

\maketitle

\begin{abstract}
 Recent work in continual learning has highlighted the beneficial effect of resampling weights in the last layer of a neural network (``zapping"). Although empirical results demonstrate the effectiveness of this approach, the underlying mechanisms that drive these improvements remain unclear. In this work, we investigate in detail the pattern of learning and forgetting that take place inside a convolutional neural network when trained in challenging settings such as continual learning and few-shot transfer learning, with handwritten characters and natural images. Our experiments show that models that have undergone zapping during training more quickly recover from the shock of transferring to a new domain. Furthermore, to better observe the effect of continual learning in a multi-task setting we measure how each individual task is affected. This shows that, not only zapping, but the choice of optimizer can also deeply affect the dynamics of learning and forgetting, causing complex patterns of synergy/interference between tasks to emerge when the model learns sequentially at transfer time.
\end{abstract}

\section{Introduction}

Despite the popularity of deep learning, neural network training is still largely considered a ``dark art" \citep{lee2020demystifying}. This alchemical connotation is in no small part due to the difficulty of building reliable intuitions about optimization in high-dimensional spaces.
Analysis of neural network loss landscapes \cite{li2018visualizing}---the map of a network's weights to their corresponding loss values---reveals which training mechanisms are effective and helps develop new methods that account for the landscape structure. 

Our work builds upon the work of \cite{javed2019metalearningrepresentationscontinuallearning,beaulieu2020learningcontinuallylearn,frati2024resetecai} and sheds new light on the effect of resampling weights during pre-training, and the dynamics of learning and forgetting while navigating the complex loss landscapes of transfer \citep{zhuang2020comprehensive} and continual \citep{continuallearningsurvey} learning problems.

What happens when \textit{zapping}---the repeated resampling of weights in the final fully connected layer of a neural network---forces the model to take sudden, sizable steps in a manifold of its weight space? The new trajectory a zapped model follows while navigating its loss landscape can surprisingly lead to better and more robust models.
When a model is transferred to a new domain it is common practice to resample\footnote{Not to be confused with \textit{resetting}, which instead restores weights to their value at initialization.} the last fully connected layer. While a random initialization helps networks learn effectively, it has been observed that this procedure can lead to a degradation of the features learned by lower layers \citep{kumar2022finetuningdistortpretrainedfeatures}, creating a ``transfer-shock" in the model. Previous work has proposed to repeatedly expose the models to similar transfer-shocks during pretraining \citep{frati2024resetecai} as a way to promote the discovery of more transferable features, and introduce noise in a model's weights which has been shown to prevent loss of plasticity \citep{dohare2024loss}.

Achieving effective transfer is particularly important for continual learning, where a model is expected to continue incorporating new knowledge, effectively experiencing constant transfer to new domains. While humans display a remarkable capacity for lifelong learning, our understanding of the mechanisms that enable continual knowledge acquisition remains limited \citep{mcclelland1995there,kumaran2016learning}. To better understand what challenges models face during sequential learning we compare the effect of different optimizers, revealing surprising behaviors at odds with the common belief that SGD outperforms adaptive optimizers in continual learning \citep[\S 4.1]{mirzadeh2020understanding}.

Our contributions are as follows:
\begin{itemize}
    \item We show the effectiveness of zapping during few-shot transfer learning on the Omni-image dataset (Fig. \ref{fig:omni-image-iid}).
    \item We show that models that have undergone zapping during training, after a transfer-shock, more easily recover compared to a non-zapped control (Fig. \ref{fig:cosim_compare}).
    \item We measure per-task losses in a challenging \textit{sequential} learning setting to reveal fine-grained patterns of learning and forgetting (\S \ref{sec:res:spaghetti}).
    \item We show that the trajectory taken by the Adam optimizer achieves continued learning and minimal forgetting of previously seen tasks (Fig. \ref{fig:frozen_1epoch}).
\end{itemize}

The combined effects of zapping and Adam result in a model which naturally maintains its prior functional form while still being updated with new knowledge.

\section{Methods}

We build upon the pre-training and continual transfer learning framework established in \cite{javed2019metalearningrepresentationscontinuallearning, beaulieu2020learningcontinuallylearn, frati2024resetecai}. In the following section we present the key mechanisms of various algorithms we compare in \S \ref{sec:res:transfer}, namely ASB, Meta-ASB and zapped-IID. For further details we provide references to algorithms and figures from previous work.

\subsection{Training Phase(s) \label{sec:met:pretrain}}
Our experimental setup consists of two primary stages: pre-training on a few-shot dataset, followed by evaluation through transfer learning of previously unseen classes.
The training phase, originally proposed in \cite{javed2019metalearningrepresentationscontinuallearning}, involves two main phases: \textit{pre-training} then \textit{transfer}.
Following \cite{frati2024resetecai}, during pre-training, models are trained using one of two possible \textit{zapping} modalities: the \textit{zapped-IID} (full-layer zap) or \textit{Alternating Sequential and Batch (ASB)} learning procedure (single-class zap) training.

When trained in zapped-IID mode the model is pre-trained using i.i.d.\footnote{To help clarify intent we use \textquote{IID} when referring to a model's pre-training (e.g. ASB vs zapped-IID) and \textquote{i.i.d.} when talking about data (e.g. i.i.d. transfer vs sequential transfer).} batches of training data while resampling the weights of the entire last fully-connected layer at the end of every epoch. This amount and frequency of zapping has been found empirically effective \citep[Tables 7 \& 8]{frati2024resetecai}.

Alternatively, the ASB mode (see \citet[\S 2.1.1 \& Alg. 1]{frati2024resetecai}) alternates between:
\begin{itemize}
    \item Sequential Learning: A class is randomly selected from the ones available in the dataset. All weights connected to that class' neuron in the final layer are resampled  (thus the class is forgotten). The model performs a series of forward passes and SGD updates on each individual example from the selected class. This sub-phase promotes fast learning of class-specific knowledge.
    \item Batch Learning: The model processes a batch of randomly sampled examples from all pre-training classes, in addition to the examples used during the Sequential Learning. This sub-phase promotes the consolidation of newly learned information, in a way that is compatible with other classes in the dataset.
\end{itemize}
 
Compared to zapped-IID, the zapping happens on a smaller scale (single class resampling vs.\ full layer) but more frequently (after every batch vs.\ every epoch end).

The model's performance can be further improved using second-order gradients. When using second-order gradients in ASB training mode, the model backpropagates the loss from the Batch Learning sub-phase through the Sequential Learning sub-phase. This meta-learning process determines the initial weights for the next ASB iteration and is inspired by the meta-learning procedure MAML, introduced in \cite{finn2017modelagnosticmetalearning} then implemented as ANML in \cite{beaulieu2020learningcontinuallylearn}, and later ablated by \cite{frati2024resetecai}, calling it Meta-ASB. See \cite[Fig. 9]{frati2024resetecai} for a visual representation of ASB and Meta-ASB.

Once a candidate model has been pre-trained, we evaluate it through transfer to novel classes. The transfer can be either: \textit{i.i.d. transfer}, where the model is now trained on batches of data from unseen classes, or \textit{continual transfer}, where we present examples to the model one at a time (\textit{sequential} learning). In both cases the datasets contain only few examples per class (\textit{few-shot} learning).

We focus on two complementary datasets that share a similar few-shot structure but differ in visual complexity:
(1) Omniglot \citep{lake2015human} provides 1,623 handwritten character classes with 20 examples per class. Its large number of classes enables extensive evaluation of catastrophic forgetting in continual learning scenarios. (2) Omni-image \citep{frati2023omnimage} contains 1,000 natural image classes with 20 examples per class, selected to maximize within-class visual consistency. This dataset maintains Omniglot's few-shot structure while introducing the complexities of natural imagery.
Across both datasets, we use 15 training examples per class during training, with the remaining examples reserved for validation (during pre-training) or testing (at transfer time).

\subsection{Zap-divergence \label{sec:met:zap-div}}

What happens to models after they get zapped during pre-training, do they get pushed onto a new learning trajectory or do they return towards their pre-zap direction?

To answer this question we employ the following protocol: we make two copies of a trained model and perturb the weights of one of the copies (treatment) while leaving the other unchanged (control). The two models are then trained on the same series of batches of data, and the similarity between layers in the treatment and control is measured at each step, thus generating a series of datapoints of shape $(\#\text{Layers}, \#\text{Steps})$, that we can use to investigate whether models further diverge or converge back towards similar weights after this initial \textquote{shock}. See Fig. \ref{fig:cosim_explained} for a visual representation of this process.

Similar to \cite{jin2020does} we use cosine similarity ($\text{cosim}(A,B) = \frac{A \cdot B}{\norm{A}\norm{B}}$) to compare the weights of different models, because of its low computational complexity---we are going to compute this similarity after every weight update. However, differently from \cite[Def. 3.1]{jin2020does} we compute the similarity between entire weight vectors instead of averaging it over individual neurons.

This protocol aims to quantify how much the training trajectory of a model diverges from or re-aligns with its counterfactual unperturbed self, due to the resampling of weights in the last layer. We refer to this protocol as \textit{zap-divergence}, and we are going to compare how zap-divergence differs for models that have undergone zapping during training compared to models who have not. Note that in Fig. \ref{fig:cosim_compare} dashed/solid lines show if a model has undergone zapping during pre-training, but both styles compare a \textit{recently} zapped model (treament) with an unperturbed version of the same model (control). This means that  solid lines indicate models that just received their first zap (i.e. a scenario similar to a sudden transfer), while dashed lines correspond to models that had received several zaps already.

\subsection{Continual per-task losses}
When transferring a trained model to a new domain it is common practice to resample the last layer's weights before resuming training \citep{yosinski2014transferable}. If only the resampled weights in the last layer are allowed to change thereafter with the remainder of the model frozen, it is also referred to as linear probing \citep{alain2018understandingintermediatelayersusing}, but doing so limits the expressivity of the network. In \S\ref{sec:res:spaghetti} we investigate both linear probing and full-model training. 
While the loss landscape is determined by the model weights, architecture and dataset, the path taken through it during training depends on the optimizer used. Adaptive optimizers are not often used in continual learning \citep{mirzadeh2020understanding} where SGD is favored, but previous work in our setting has shown Adam to be an effective choice. See Appendix \ref{appendix:optim} for a comparison of the SGD and Adam algorithms.

To better understand the dynamic of learning and forgetting in continual transfer learning with different optimizers, in \S \ref{sec:res:spaghetti} we are going track \textit{separately} the loss for each of 100 tasks, over multiple epochs. The resulting plots contain 100 lines, color coded by training order from first to last (as cooler to warmer: \vbars).
By focusing on the per-task loss plots we more accurately highlight how different tasks interfere or synergize during training, represented by loss increase or continued improvement on tasks not currently trained on. We randomize the order of classes in the dataset to break any possible correlation between examples (e.g. characters within the same alphabet), but keep the random order consistent across multiple replicates.

\section{Results}

We begin our investigation by exploring performance on the challenging Omni-image dataset (\S\ref{sec:res:transfer}), comparing the effect of several pre-training strategies (IID, ASB, and Meta-ASB), with and without zapping, in a standard i.i.d.\ transfer setting. In \S\ref{sec:res:zap-div}, motivated by the effectiveness of zapped-IID as a pre-training method, we investigate the effect of zapping during pre-training from the perspective of layer re-alignment. Finally, in \S\ref{sec:res:spaghetti}, we take a close look at the pattern of learning and forgetting during sequential learning by tracking each individual task loss separately, revealing the complex dynamics that depend on the choice of optimizer. To reduce the computational cost of these investigations, we focus on the Omniglot dataset in \S\ref{sec:res:zap-div} and \S\ref{sec:res:spaghetti}.

\subsection{Transfer Learning \label{sec:res:transfer}}

\begin{figure}[!htbp]
    \centering
    \begin{minipage}{0.50\textwidth}
        \centering
        \includegraphics[width=\linewidth]{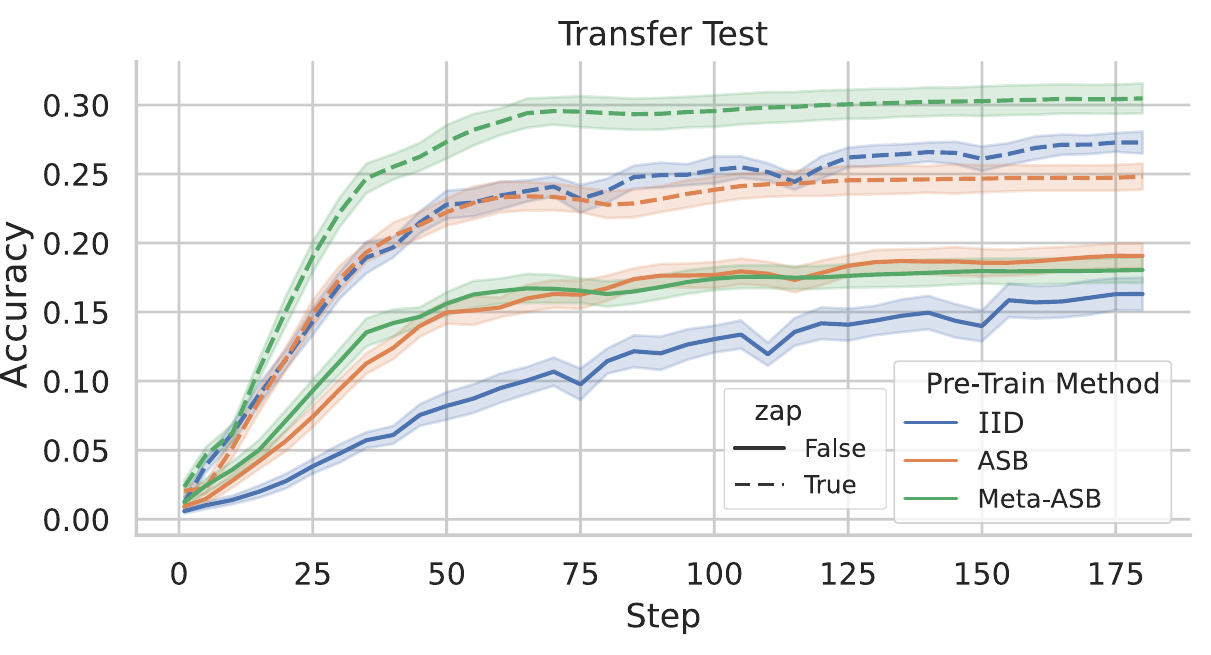}
        \caption{Test accuracy on classes during standard fine-tuning on the omni-image subset of ImageNet (15 training images / 5 test images per class). Models pre-trained \textbf{with zapping} achieve significantly higher test accuracies, with models employing both meta-gradients and zapping coming out on top.}
        \label{fig:omni-image-iid}
    \end{minipage}
    \hfill
    \begin{minipage}{0.48\textwidth}
        \centering
        \begin{tabular}{l@{\hspace{2pt}}l@{\hspace{6pt}}c@{\hspace{2pt}}c}
    \toprule
    \multicolumn{2}{c}{\bf Pre-Train Method}
    &\multicolumn{1}{c}{\bf Pre-Train}
    &\multicolumn{1}{c}{\bf Transfer Test} \\
    \midrule
    \solidline{C0} & IID & 27.2 {\scriptsize $\pm 0.5$} & 16.3 {\scriptsize $\pm 1.2$} \\
    \rowcolor{mygray}
    \dashedline{C0} & IID+zap & 26.8 {\scriptsize $\pm 2.1$} & \textbf{27.3} {\scriptsize $\pm 0.8$} \\
    \solidline{C1} & ASB & 18.6 {\scriptsize $\pm 0.3$} & 19.1 {\scriptsize $\pm 0.9$} \\
    \rowcolor{mygray} 
    \dashedline{C1} & ASB+zap & 20.9 {\scriptsize $\pm 1.3$} & \textbf{24.8} {\scriptsize $\pm 0.9$} \\
    \solidline{C2} & Meta-ASB & 22.0 {\scriptsize $\pm 1.0$} & 18.0 {\scriptsize $\pm 0.9$} \\
    \rowcolor{mygray} 
    \dashedline{C2} & Meta-ASB+zap & 28.9 {\scriptsize $\pm 0.7$} & \textbf{30.5} {\scriptsize $\pm 1.1$} \\
    \bottomrule
\end{tabular}
        \captionof{table}{Average accuracy ($\pm$ std dev) for the standard fine-tuning transfer problem. 
        \textbf{Pre-Train} is the final validation accuracy of the model on the \textit{pre-training} dataset.
        \textbf{Transfer} is the accuracy on held-out instances from the \textit{transfer-to} dataset after five epochs of fine-tuning.
        }
        \label{tab:full_iid}
    \end{minipage}
\end{figure}

Previous work investigating the effect of zapping has shown its effectiveness on the Omniglot dataset but due the simplicity of tasks used (i.e. handwritten characters) all methods achieved very similar accuracy \citep[Fig. 5a]{frati2024resetecai}.
Since the Omni-image dataset retains the same structure as Omniglot but uses more complex natural images we can now more clearly separate the effect of individual treatments (Meta-ASB vs ASB, zap vs no-zap).

Comparison of results between pretraining and transfer shows that zapped models not only outperform their non-zapped counterparts at transfer time (e.g. $16.3\%$ for models pretrained on i.i.d. data without zapping, compared to $27.3\%$ with zapping) but also improve upon their final validation accuracy before transfer (e.g., $20.9\%$ without zapping during ABS pretraining, compared to $24.8\%$ with zapping). This shows that zapping pretraining allows the model to learn effectively at transfer time.
We also observe that the final test performance of IID (27.3\%) is slightly but significantly ($p=0.0017$) higher than that of ASB (24.8\%).

Without zapping (solid lines), ASB learn significantly faster than IID (Fig.\ref{fig:omni-image-iid} epochs 0-50), but with zapping (dashed lines) their speed of improvement on the test set is the same.
  
While using higher-order gradients (Meta-ASB) provides the best performance, the computational cost increases significantly compared to IID. On the other hand, the relatively inexpensive zapped-IID setting approaches the test performance of Meta-ASB ($27.3\%$ and $30.5\%$ respectively) without additional compute and slightly improves on the more complex ASB algorithm ($24.8\%$).

Overall these results further show the essential contribution of zapping, and in the next section, we focus our attention on the zapped-IID mechanism to better understand its effect during pre-training.

\subsection{Zap-divergence \label{sec:res:zap-div}}

As described in \S\ref{sec:met:pretrain}, the IID approach resamples the entire last layer, which aligns with the common practice of replacing the last layer before transfer learning. Prior work from \cite{kumar2022finetuningdistortpretrainedfeatures} demonstrated that last-layer resampling can distort learned features and degrade network performance.

This presents an apparent contradiction: while zapping typically impairs performance during transfer learning, zapping during training yields consistently positive results across datasets. To investigate this phenomenon, we measure the zap-divergence metric (\S \ref{sec:met:zap-div}) on the Omniglot dataset to analyze how networks which have been previously zapped during pretraining respond, compared to models who have not being previously zapped.

\begin{figure}[!htb]
    \centering
\vspace{-0.1in}
\begin{subfigure}[t]{0.3\textwidth}
    \includegraphics[width=\linewidth]{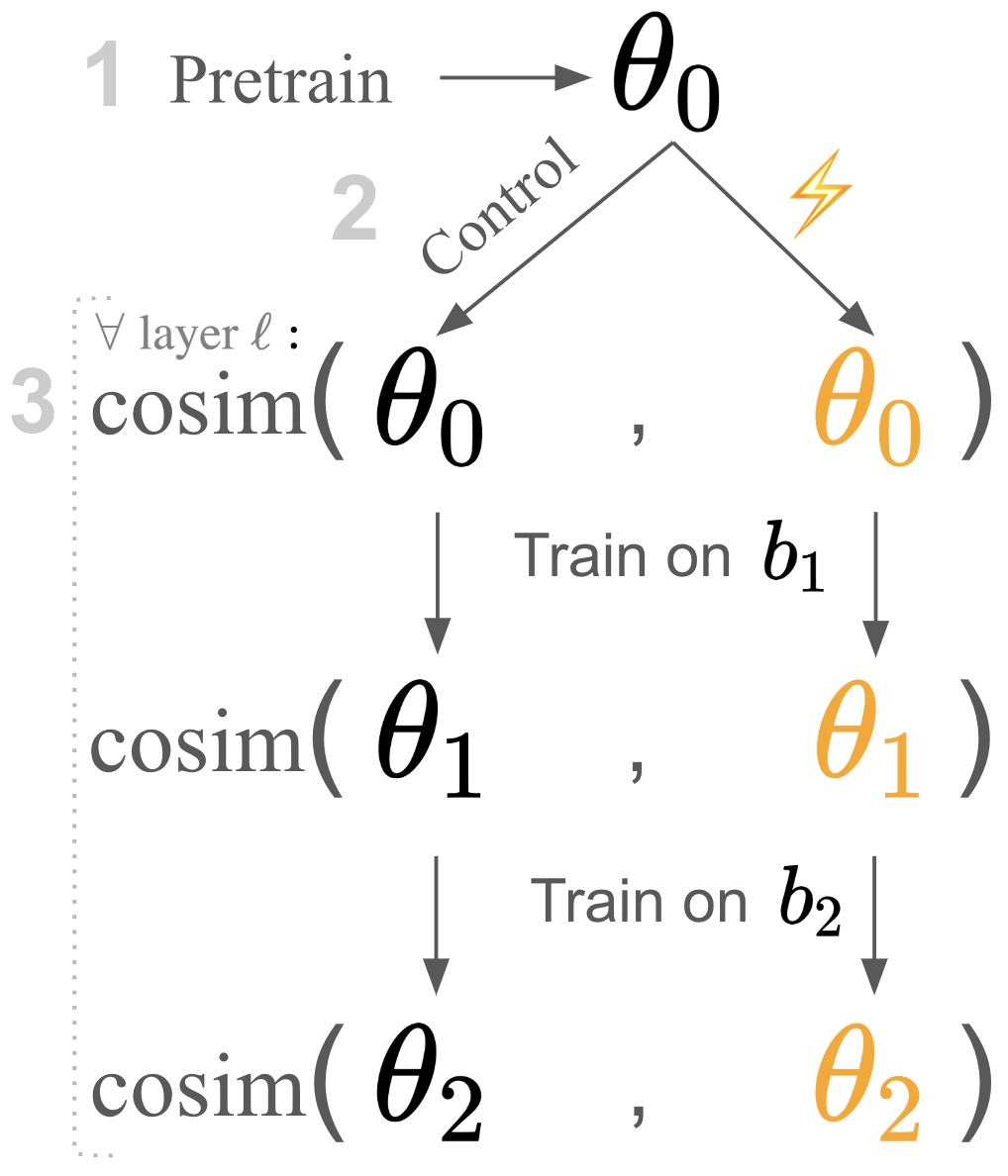}
    \caption{ }%
    \label{fig:cosim_explained}
\end{subfigure}
\begin{subfigure}[t]{0.65\textwidth}
    \centering
    \includegraphics[width=\linewidth]{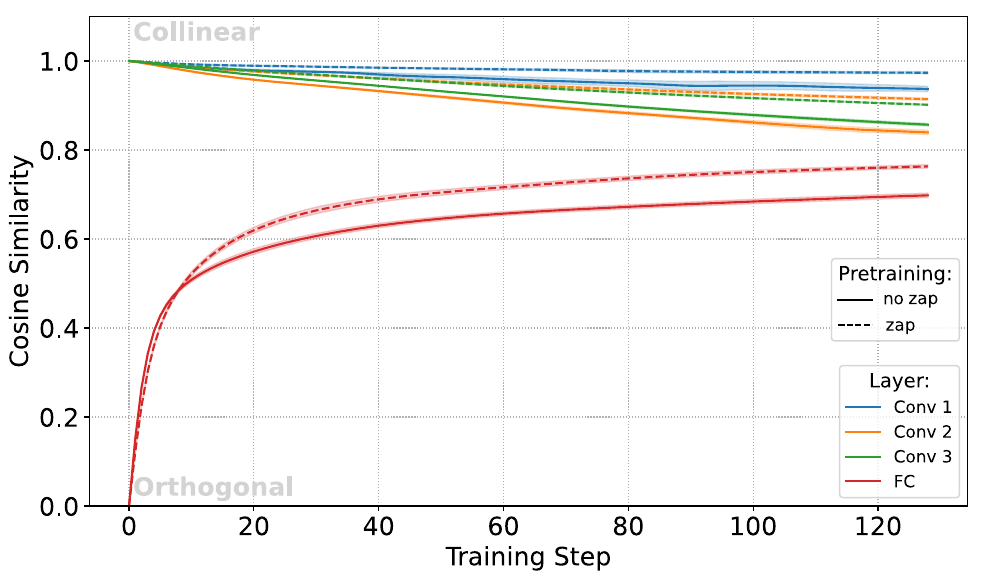}
    \caption{ }%
    \label{fig:cosim_compare}
\end{subfigure}
\caption{Measuring the effect of zapping the last layer of a model during i.i.d. training with the Adam optimizer. \textbf{(a)} Measuring the robustness to perturbations of a training trajectory: (1) Starting with a pretrained model with parameters $\theta_0$, (2) we create two variants - a \textit{control} model that maintains $\theta_0$ (left) and a \textit{perturbed} model where the last layer has been resampled $\theta_0$\raisebox{0.5ex}{\Lightning} (right) once before training. (3) Both models are then trained on the same mini-batches $b_1, b_2,\cdots$. We measure the cosine similarity between layers $\ell$ in the \textit{control} and \textit{perturbed} models after each batch and record this similarity, shown in Fig. \ref{fig:cosim_compare}. \textbf{(b)} The last \textcolor{C3}{fully connected (FC)} layer is re-sampled and then the whole model is trained on a series of i.i.d. minibatches as explained in Figure \ref{fig:cosim_explained}. We observe that \textcolor{C0}{Conv1}, \textcolor{C1}{Conv2}, \textcolor{C2}{Conv3} are much less affected when the model has undergone zapping during training (e.g. \solidline{C0} vs \dashedline{C0}) and, that the \textcolor{C3}{FC} layer returns to values that are more similar to the control that didn't undergo re-sampling at the beginning of this training trajectory.}
\end{figure}

Figure \ref{fig:cosim_compare} illustrates the zap-divergence for a network trained in the IID setting, with dashed lines representing the zapped version and solid lines representing the non-zapped version. 

As training progresses, we observe that:
\begin{enumerate}
    \item After experiencing a perturbation meant to simulate a transfer shock, if a model was pretrained with zapping, its FC layer more easily re-aligns with its unperturbed control (Fig.\ref{fig:cosim_compare}, dashed red is above solid red). This may suggest that the FC layer progressively converges on a more stable state over the course of repeated zappings.
    \item Even though the conv layers were not directly perturbed, the effect of the shock in the FC layer flows down into those layers during subsequent training steps. If a model was pretrained with zapping, these lower layers are less affected by the last layer perturbation than if the model had not been subjected to zapping during pretraining.
\end{enumerate}

Together these findings show that during training, the network has managed to find a weight configuration that allows both a faster recovery from (in the FC layer) and is less affected by (in lower layers) the ``transfer-shock" caused by resampling the last layer.

\begin{figure}[!htb]
    \centering
    \begin{subfigure}[t]{0.4\textwidth}
        \centering
\includegraphics[width=\linewidth]{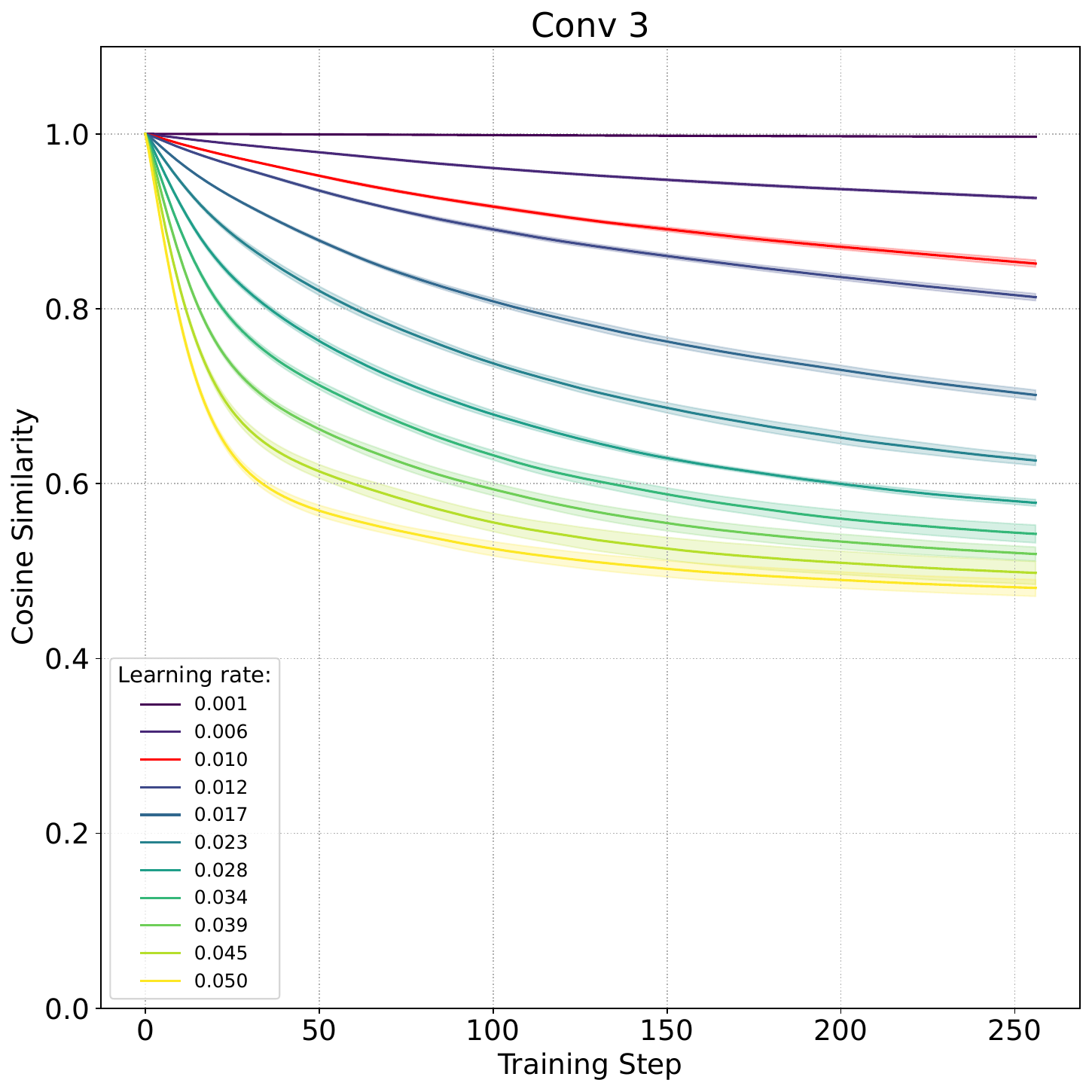}
    \caption{Conv 3 layer: as the learning rate increases (\vbars) the zapped version quickly diverges from the control, until it reaches a similarity of 0.5. Across all learning rates the behavior is one of monotonic decrease, as opposed to that observed in the FC layer.}
    \label{fig:cosim_encoder2}
    \end{subfigure}
    ~
    ~
    ~
    \begin{subfigure}[t]{0.4\textwidth}
         \centering
\includegraphics[width=\linewidth]{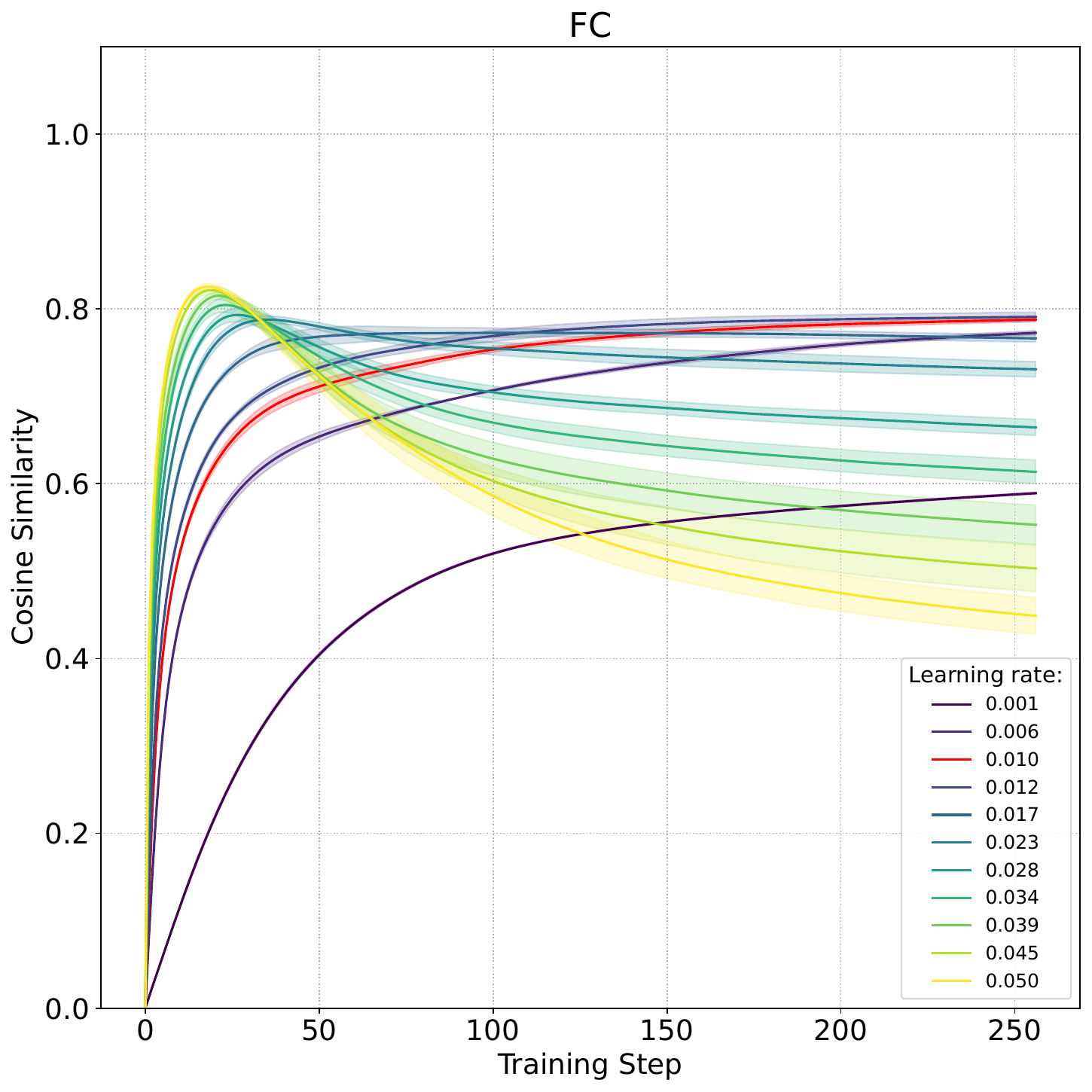}
    \caption{FC layer: as the learning rate increases (\vbars) the zapped version quickly moves away from orthogonality (i.e. cosim. $> 0$) across all learning rates. However, at higher learning rates the initial approach phase ($\text{cosim.}\uparrow$) is followed but a rapid divergence phase ($\text{cosim.}\downarrow$)}
    \label{fig:cosim_fc}
    \end{subfigure}
    \caption{Cosine similarity of layers after re-sampling the FC layer, using the Adam optimizer across 10 learning rates (we highlight in red \solidline{red} the learning rate used in Fig. \ref{fig:cosim_compare}).}
    \label{fig:cosim_lrs}
\end{figure}

Results in Fig. \ref{fig:cosim_compare} have been obtained using the default learning rate used in most of the experiments. However, the amount of learning and forgetting, and the rate of re-alignment after a zap depends on the learning rate used. In Fig. \ref{fig:cosim_lrs} we show how the re-alignment changes as we sweep over a range of 10 learning rates (see Appendix Fig. \ref{app:fig:all_cosims} for all layers).

In the last convolutional layer (Fig. \ref{fig:cosim_encoder2}) we see that the alignment starts at $1$ (since this layer has not been resampled), and then decreases as the training continues. As expected, the decrease in alignment is faster for higher learning rates, but interestingly it plateaus. In the fully-connected layer (Fig. \ref{fig:cosim_fc}) the alignment starts at zero since the weights have been resampled and quickly increases during training. We observe that, while no treatment reaches an alignment above $0.8$, higher learning rates align more quickly but then diverge. Interestingly the default learning rate used in most experiments (highlighted in red) achieves a good balance of speed of recovery and final alignment.

Overall we observe that lower layers benefit from lower learning rates (to not forget previous features), while the fully connected layer needs a higher learning rate to recover quickly, but not too high to avoid diverging. This is in line with the empirical observation that these models are very sensitive to learning rates in their few-shot transfer evaluation---to learn quickly the learning rate has to be high, but to not forget the learning rate has to be low. We hypothesize that zapped models settle in wider loss basins which allows models to take more aggressive steps (higher learning rate) while remaining within the basin.

\subsection{Per-task continual losses\label{sec:res:spaghetti}}

\begin{figure}[!ht]
\centering
\begin{subfigure}[b]{0.32\linewidth}
   \centering
   \begin{tikzpicture}
       \node[anchor=south west] (img) at (0,0) {\includegraphics[width=\linewidth]{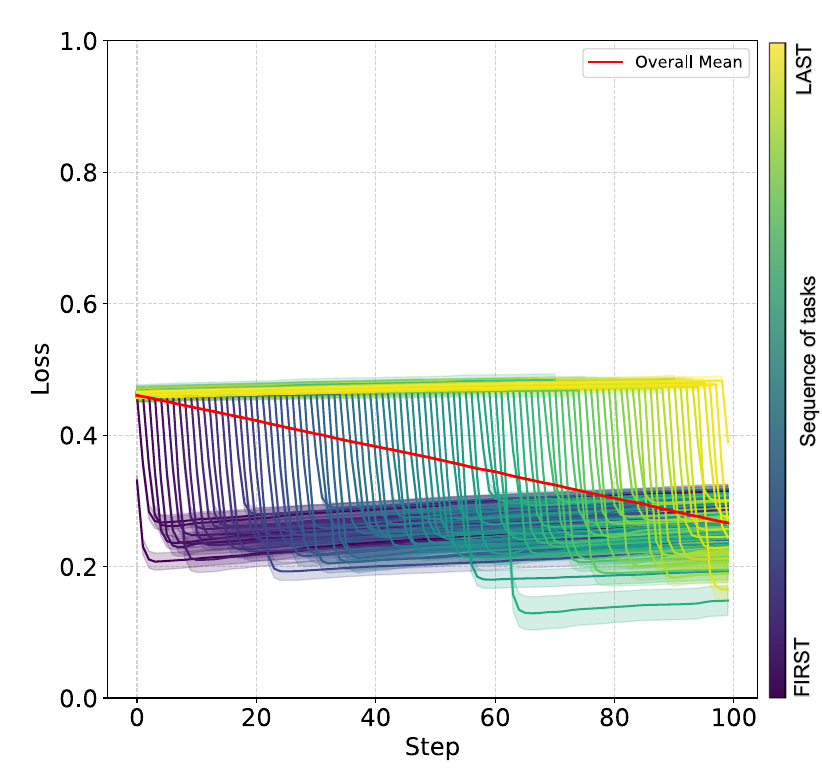}};
       \node[lightgray, font={\Large\sffamily}] at (2,1.2) {A};
       \node[lightgray, font={\Large\sffamily}] at (3.5,3) {B};
   \end{tikzpicture}
   \caption{linear probing w/ SGD\\($\gamma = 0.0010, \mu = 0.9$ )}
   \label{subfig:frozen_sgd_high}
\end{subfigure}
~
~
\begin{subfigure}[b]{0.3\linewidth}
   \centering
   \begin{tikzpicture}
       \node[anchor=south west] (img) at (0,0) {\includegraphics[width=\linewidth]{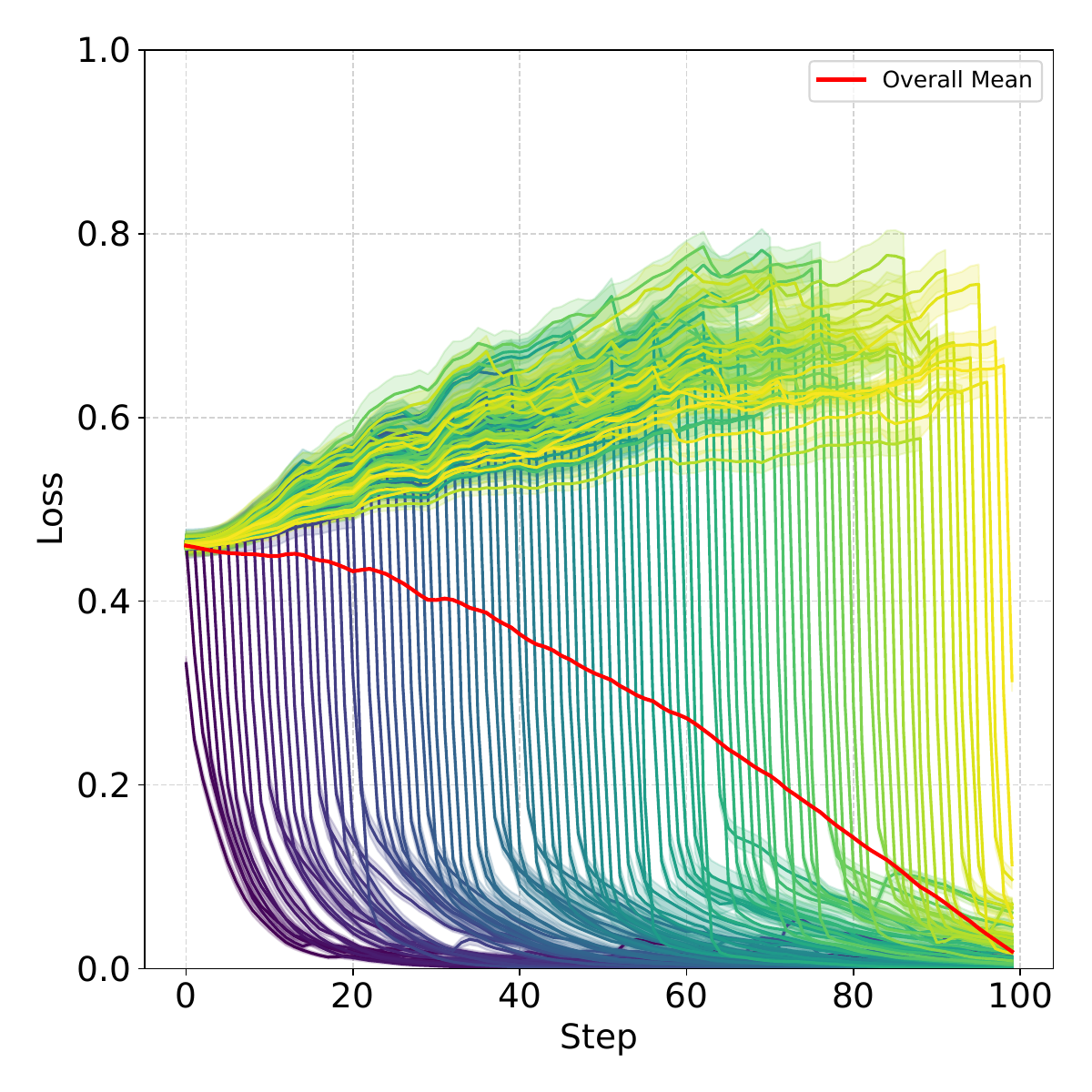}};
       \node[lightgray, font={\Large\sffamily}] at (1,1) {A};
       \node[lightgray, font={\Large\sffamily}] at (3.5,4.3) {B};
   \end{tikzpicture}
   \caption{linear probing w/ Adam\\($\gamma = 0.0010, \beta_1=0.9$)}
   \label{subfig:frozen_adam_high}
\end{subfigure}
~
~
\begin{subfigure}[b]{0.3\linewidth}
   \centering
   \begin{tikzpicture}
       \node[anchor=south west] (img) at (0,0) {\includegraphics[width=\linewidth]{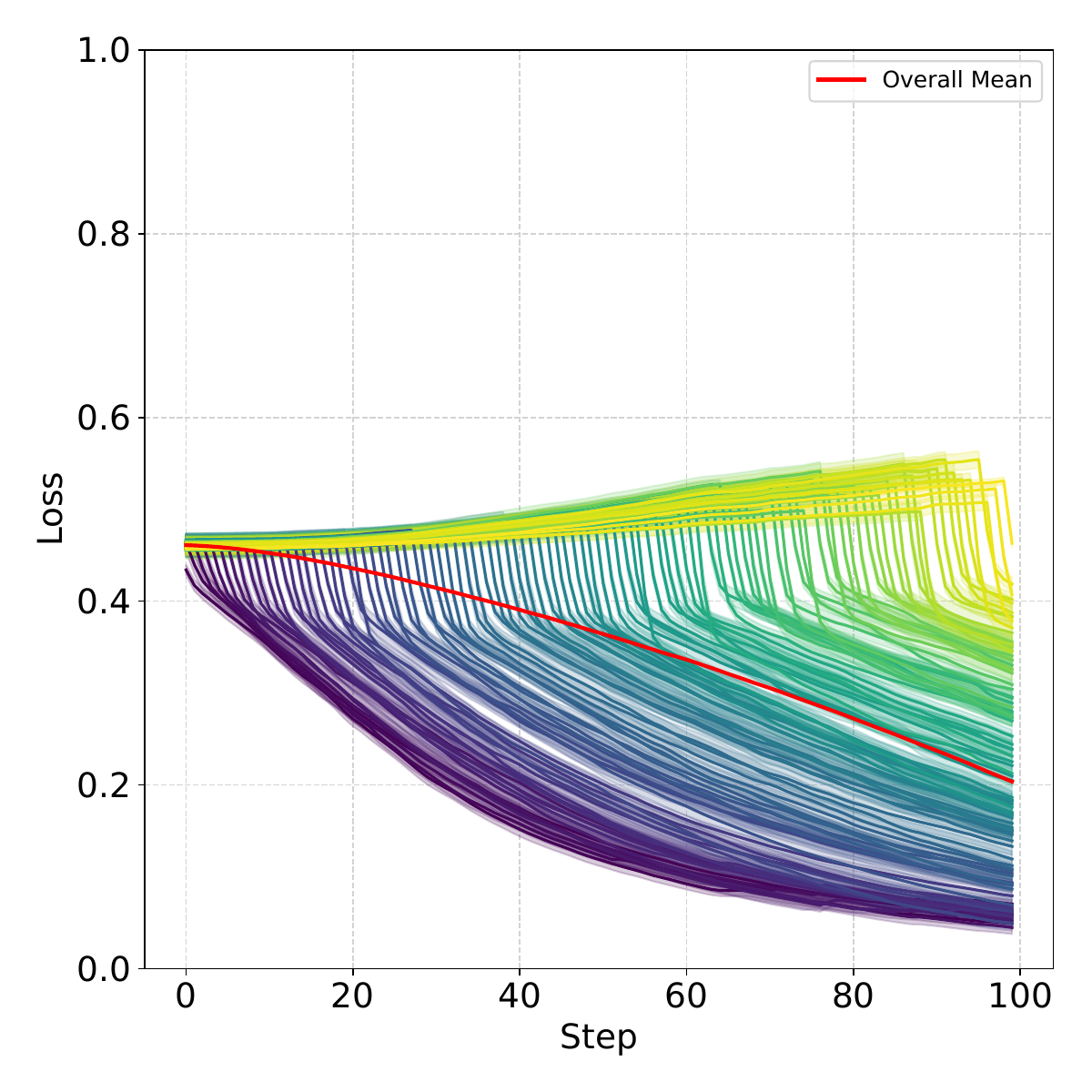}};
       \node[lightgray, font={\Large\sffamily}] at (2.0, 1.2) {A};
       \node[lightgray, font={\Large\sffamily}] at (3.5, 3.3) {B};
       \node[lightgray, font={\Large\sffamily}] at (5.1, 1.65) {C};
       \draw[lightgray, decorate, decoration={brace}, thick] (4.75,2.5) -- (4.75,0.8);
   \end{tikzpicture}
   \caption{linear probing w/ Adam\\($\gamma = 0.0002, \beta_1=0.9$)}
   \label{subfig:frozen_adam_low}
\end{subfigure}
\caption{Comparison of learning dynamics in linear probing (colors indicate order of training): Both optimizers store gradient information but Adam leads to much longer lasting improvements of individual tasks and better final performance.}
\label{fig:frozen_1epoch}
\end{figure}

Examining the zap-divergence of models across different learning rates highlights a fundamental tension between two competing requirements: the resampled layer of the model requires higher learning rates to recover quickly, while unaffected layers need lower learning rates to maintain stability. This tension parallels the balance between learning and forgetting that models experience during continual learning, where rapid acquisition of new information risks compromising previously learned tasks, also known as the ``plasticity-stability dilemma" \citep{mermillod2013stability}. Previous works \citep{beaulieu2020learningcontinuallylearn, frati2024resetecai} have used both the SGD and Adam optimizers at different stages in the training and transfer phases but have not investigated the effect of these choices in the sequential learning setting.
Here, we show that the optimizer choice can lead to learning dynamics where loss continues improving on tasks not currently trained (Fig. \ref{subfig:frozen_adam_low} \& \ref{fig:sub:unfrozen_adam_low}), and the nuanced trade-off between learning speed and final performance. The following results are obtained during sequential transfer training of zapped models. While zapping significantly improves the performance of models it only has a minor effect of the magnitude of the patterns observed (see Fig \ref{fig:zap_effect_pertask} in Appendix for a comparison). 

\paragraph{Continual linear probing.} In Fig. \ref{fig:frozen_1epoch} we show several examples of per-class losses during the first epoch of transfer training, using linear probing.  

In Fig. \ref{subfig:frozen_sgd_high} the model is trained using the SGD optimizer and the pattern of learning and forgetting is as expected, every time a class is trained the loss drops sharply and is then followed by a slow rise while training on other classes causes some degree of forgetting (see \ref{subfig:frozen_sgd_high}.\figmark{A}); losses corresponding to other classes that are not yet trained on remain unaffected (see \ref{subfig:frozen_sgd_high}.\figmark{B}). 

In Fig. \ref{subfig:frozen_adam_high} the same training is done using the Adam optimizer. First, we notice that the model experiences a much higher degree of interference on tasks that haven't been trained yet (rising sharply in \ref{subfig:frozen_adam_high}.\figmark{B}), while with SGD training one task barely affected the others (compare \ref{subfig:frozen_sgd_high}.\figmark{B} to \ref{subfig:frozen_adam_high}.\figmark{B}). However, this increased interference is more than balanced by much faster learning on the current task, which leads to overall better performance. Second, we observe that loss continues decreasing for many steps even after training on that task has finished (see slope of curves at \ref{subfig:frozen_adam_high}.\figmark{A}). This phenomenon is referred to in continual learning as ``backward transfer'' (i.e., loss improvement in a previous task while training on a later one) and is particularly noticeable when using lower learning rates (see Fig. \ref{subfig:frozen_adam_low}). This behavior is almost absent when training with SGD despite using an amount of momentum comparable to the Adam setting  (SGD.$\mu = 0.9 =$ Adam.$\beta_1$). 

The patterns we observe are a consequence of the sequential learning procedure used after transfer, where each example from each task is presented one at a time (see Fig. \ref{fig:iid_spaghetti} in the Appendix for an example of what one of these per-task plots would look like in a traditional i.i.d. training regime). Interestingly, in the sequential case with low learning rate the losses are quite evenly spaced out over the whole range (see Fig. \ref{subfig:frozen_adam_low}.\figmark{C}), showing that as more tasks are added the learning on previous ones continues at roughly the same rate.

While the previous plots were investigating the learning dynamic of the linear probing setting, we conclude our investigation by looking at the structure of per-task losses during full-model training.

\begin{figure}[!htb]
    \centering
    \begin{tikzpicture}
       \node[anchor=south west] (img) at (0,0) {\includegraphics[width=\linewidth]{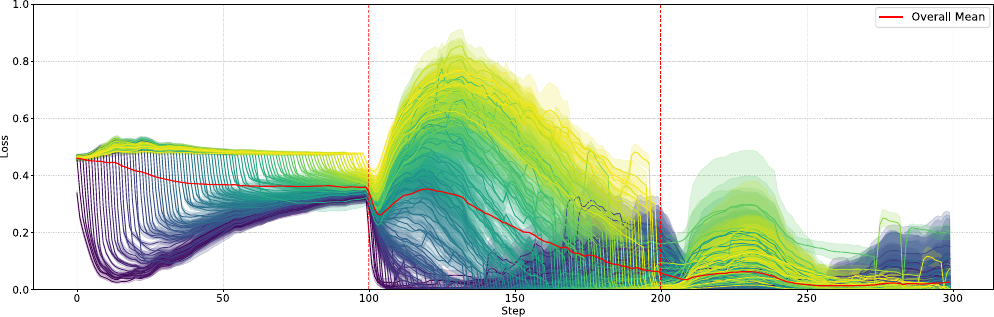}};
       \node[lightgray, font={\Large\sffamily}] at (4.0, 1.2) {A};
       \node[lightgray, font={\Large\sffamily}] at (5.5, 3.3) {B};
       \node[lightgray, font={\Large\sffamily}] at (7.0, 2.8) {C};
       \node[lightgray, font={\Large\sffamily}] at (10., 3.6) {D};
       \node[lightgray, font={\Large\sffamily}] at (13., 3.0) {E};
   \end{tikzpicture}
    \caption{Full-model sequential training using Adam ($\gamma=0.0008, \beta_1 = 0.9$) on Omniglot, 100 tasks. Letters \figmark{A}-\figmark{E} correspond to phenomenon discussed in the text. Vertical dashed red lines indicate the end of an epoch.}
    \label{fig:unfrozen_adam}
\end{figure}

\paragraph{Continual full-model tuning.} Full-model sequential training with the Adam optimizer (Fig. \ref{fig:unfrozen_adam}) is remarkably different than during linear probing.
Here we observe that: (1) tasks show a high degree of interference as loss for a given task worsens after training on it ceases (Fig.\ref{fig:unfrozen_adam}.\figmark{A}); (2) tasks don't worsen \textit{before} training them (Fig.\ref{fig:unfrozen_adam}.\figmark{B}) but the improvement in loss per task shrinks for each subsequent task (compare first and last tasks trained); (3) tasks show a huge amount of interference during the second epoch inversely correlated to the order of training (Fig.\ref{fig:unfrozen_adam}.\figmark{C}; note that tasks are fixed in a shuffled order so consecutive ones are not coming from the same alphabets or would otherwise expect to be correlated); (4) the interference at the beginning of the second epoch reverts during the course of the second epoch even before the affected tasks are trained again (Fig.\ref{fig:unfrozen_adam}.\figmark{D}) and results in some tasks improving drastically between the time they finish training in the first epoch (Fig. \ref{fig:unfrozen_adam}.\figmark{B}) and when they begin training in the second epoch; (5) following this, the amount of interference diminishes in subsequent epochs (Fig. \ref{fig:unfrozen_adam}.\figmark{E} is less pronounced than Fig. \ref{fig:unfrozen_adam}.\figmark{C}).
See Fig. \ref{fig:unfrozen_adam_comparison} for a comparison of different learning rates. 
Overall, it's unclear why these surprising dynamics occur, and studying them is an interesting direction for future work.

\begin{figure}[!htb]
    \centering
    \includegraphics[width=\linewidth]{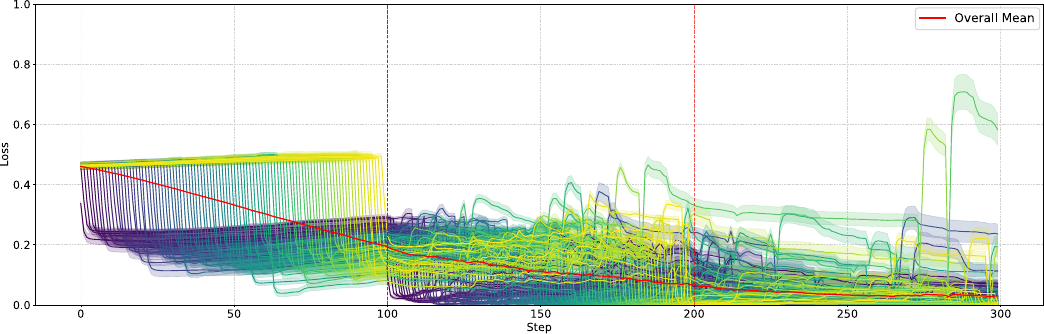}
    \label{subfig:unfrozen_sgd}
\caption{Full-model sequential training using SGD ($\gamma=0.0008, \mu = 0.9$) on Omniglot, 100 tasks.Vertical dashed red lines indicate the end of an epoch, .}
\label{fig:unfrozen_sgd}
\end{figure}

When using SGD, full-model sequential training (Fig. \ref{fig:unfrozen_sgd}) shows a behavior qualitatively similar to the linear probing case but as learning continues the loss becomes more chaotic with sudden loss spikes, indicating interference between learned tasks. If we compare the accuracies achieved by both optimizers in the continual full-model training setting, we see that SGD learns much more effectively in the first epoch (Accuracy $\text{SGD}=65.5\% > 28.0\%=\text{Adam}$) but Adam achieves higher accuracy with 2 more epochs (Accuracy $\text{Adam}=85.8\% > 76.7\%=\text{SGD}$). See Table \ref{tab:frozen_acc_10reps} \& \ref{tab:unfrozen_acc_10reps} in Appendix for all the accuracy values in both the linear probing and full-model settings.

To recap, in this section we have shown that:
\begin{itemize}
\item During sequential learning when using Adam with low learning rates, loss can continue decreasing even after training on a specific task has ended, and momentum alone does not explain this phenomenon (see Fig. \ref{fig:frozen_1epoch}).
\item Because of the aforementioned behavior requires low learning rates, Adam can be outperformed by SGD during initial epochs but surpass SGD performance in later epochs (see Table \ref{tab:unfrozen_acc_10reps}).
\item Analyzing the per-task losses of sequential training with Adam reveals surprising patterns of synergy and interference during training (see Fig. \ref{fig:unfrozen_adam}) that highlight gaps in our understanding of the dynamics underlying sequential learning. 
\end{itemize}

\section{Related Work \& Discussion}
\subsection{Plasticity, zapping and transfer learning}
As training of a deep neural network progresses the accuracy increases, but plasticity---the ability to quickly adjust outputs in response to new information---decreases.
Maintaining the plasticity of a neural network could enable it to adapt to new tasks even after extensive training, a property that is crucial for lifelong learning in dynamic environments. Some approaches, inspired by neuroscience, leverage mechanisms such as Hebbian plasticity \citep{miconi2018differentiable, najarro2020meta}, which emulate biologically plausible learning processes. Alternatively, other studies \citep{nikishin2022primacy, lyle2023understanding, chen2023improving, dohare2024loss} have explored the impact of resetting or resampling weights during training—--a simple yet effective strategy to restore the high plasticity observed in networks initialized with random weights. These methods highlight promising directions for maintaining adaptability in neural networks, even after extensive training.

While the standard approach when transferring a trained model to a new domain has been to just resample and retrain the last layer, \cite{yosinski2014transferable, kumar2022finetuningdistortpretrainedfeatures} show that this approach incurs a trade-off. If most of the model is kept frozen the effectiveness of the fine-tuning is reduced. If the whole model is allowed to update, fine-tuning can disrupt fragile co-adapted features and distort previously learned tasks. Exposing a model to ``transfer shocks" during training can promote the discovery of more adaptable features. Rehearsing noisy events has also proved useful when introducing noise at the sub-parameter level through quantization.

To improve the computational efficiency of networks, it is common practice to reduce the precision of weights used in models, but the computation savings come at the cost of added noise from the reduced precision. Rather than reducing the precision of the weights only once after training has been completed, quantization-aware training (QAT) has become the de facto standard \citep{hubara2018quantized, krishnamoorthi2018quantizing}. By simulating the quantization operation during training, the network can adapt to the quantization noise. It stands to reason that similar improvements should be observed by simulating the noise induced by resampling weights during transfer, effectively implementing a transfer-aware training regime by repeatedly zapping the model.

\subsection{Why does zapping help learning? escape early lock-in\label{lit:jumps}}

Work investigating the parameter space of neural networks has shed light on the structure of loss basins---regions in parameter space where points achieve similarly low loss--- that neural networks traverse during training. These loss basins are both quite high dimensional \citep{li2018measuringintrinsicdimensionobjective}, and connected to each other through geometrically simple, but hard to find, paths \citep{garipov2018losssurfacesmodeconnectivity}.

If the loss landscape of neural network training contains large \citep{li2018measuringintrinsicdimensionobjective} connected \citep{garipov2018losssurfacesmodeconnectivity, draxler2018essentially} basins, why is training so challenging in practice?
Neural networks can be studied from two perspectives: the parameter space, where each set of parameters is mapped to a loss value; or the function space, which focuses on the predictive distribution described by the network. A limitation of the parameter space point of view is that many different functions may achieve the same loss, thus the parameter space cannot tease apart the details of the predictions. Alternatively, the function space point of view shows that predictions can vary across different basins \citep[Fig. 5]{fort2020deepensembleslosslandscape}. This allows us to see how easy-to-find basins can lock models into a particular functional form early on in training (\citet[Fig. 7, left]{fort2020deeplearningversuskernel}; \citet[Fig. 4]{fort2020deepensembleslosslandscape}). While it is known that weight changes early on in training are crucial for final performance, which features of those changes \cite[\S 5]{frankle2020earlyphaseneuralnetwork} are most important remains unclear. From this point of view, \textit{zapping} (i.e. forget and re-learn) of classes (ASB) or layers (zapped-IID) provides a learning-dynamic-driven way to encourage exploration of the function space, by escaping early lock-in.

\subsection{Adam and continual learning: potential connections with the Fisher Information Matrix and Natural Gradient Descent \label{lit:adam}}

We observe that in our sequential learning setup, model performance on past tasks can continue to improve even after training on those specific tasks concludes. This implies the optimizer leverages stored information across task transitions. Momentum, which relies on recent gradient history, fails to fully account for this sustained improvement on prior tasks (Fig. \ref{fig:frozen_1epoch}). In contrast, the Adam optimizer also tracks squared gradients (second-moment information). In the following, we review literature highlighting how this second-moment information can be advantageous in a continual learning settings, and potentially relate to the observed phenomenon.

Traditional Stochastic Gradient Descent (SGD) works by iteratively optimizing the weights $\theta$ of a neural network by following the gradient of a chosen loss $\mathcal{L}$ on a batch of data.
While the gradient of the loss represents the direction of maximum local decrease of the loss, this does not guarantee that the iterative process will preserve previously acquired knowledge.
Instead of purely following the steepest descent direction, it would be beneficial if weight updates could seek a path that minimizes the loss while preserving the network's existing behavior. Natural Gradient Descent (NGD, \cite{amari1998natural}) achieves this by taking into consideration the topology of the search space, and replacing the Euclidean distance used in the parameter space with the KL divergence between predictive distributions. NGD modifies the gradient as
$\Tilde{\nabla}_\theta \mathcal{L} = F_\theta^{-1}\nabla_\theta\mathcal{L}$ where $F_\theta = \mathbb{E}_{x \sim p} \left[ \nabla_\theta \log p(x;\theta) \, \nabla_\theta \log p(x;\theta)^\top \right]$, $F$ is the Fisher Information Matrix (FIM) (see \cite{martens2020new} for more details).
Because the NGD minimizes the KL divergence between the predictive distributions before and after an update, it has led to using the FIM as an additional loss in continual learning \citep{kirkpatrick2017overcoming}.

Unfortunately, the FIM is impractical to compute (the true data distribution is unknown and sampling the posterior is expensive) and massive to store (its size is the number of parameters of the model squared). %
To address this limitation several works such as \cite{kirkpatrick2017overcoming} use only the diagonal of the empirical FIM computed from batches of data, which amounts to squaring the gradients. The Adam optimizer \citep{kingma2017adammethodstochasticoptimization} computes the EMA of squared gradients, which has been characterized as using an Empirical Fisher (EF) preconditioning. However, rather than using the squared gradients directly, Adam applies a square root transformation (Eq. \ref{eq:adam:sqrt}), to achieve invariance to gradient magnitude \citep[\S 2.1]{kingma2017adammethodstochasticoptimization}, \citep[\S 4.3]{kunstner2019limitations}. While subsequent research shows that the square root transformation is crucial---both higher and lower power values lead to degraded performance \cite[\S C.1.1]{hwang2024fadamadamnaturalgradient}---we hypothesize that the connection with the empirical FIM could contribute to the surprising capability of Adam to learn effectively in our continual learning setting. 
These findings challenge the popularity of SGD over Adam in several continual learning algorithms such as EWC \citep{kirkpatrick2017overcoming}, PackNet \citep{mallya2018packnet}, or iCarl \citep{rebuffi2017icarl}.

\section{Conclusion}

Our investigation into weight resampling (``zapping") during neural network training has revealed several key mechanisms that contribute to improved performance in challenging learning scenarios. Through detailed analysis of learning dynamics and weight space trajectories, we have shown that the benefits of zapping fundamentally alter how networks learn and adapt to new tasks.

Our experiments demonstrated that models trained with zapping show remarkable resilience to the ``transfer shock" typically associated with last-layer resampling. When the final layer is resampled, in zapped models it more quickly realigns with their pre-zap trajectories (Fig. \ref{fig:cosim_compare}), suggesting they have developed more robust and transfer-friendly feature representations in their lower layers. This finding helps explain why zapped models consistently outperform their non-zapped counterparts across different pre-training strategies (IID, ASB, and Meta-ASB).
Our analysis of optimizer dynamics revealed an important interplay between Adam's gradient memory and sequential learning.
Our investigation of full-model training in a sequential setting showed that SGD was faster at learning in the first epoch, but that the the squared-gradient information recorded by Adam allowed it to continue to improve long after a task was done training and eventually achieved significantly superior accuracy ($76.7\%$ vs $85.8\%$, see Table \ref{tab:unfrozen_acc_10reps}). This shows an interesting synergy between Adam and sequential learning and future work should focus on studying the different gradient directions tracked by adaptive optimizers, to better understand under which conditions local gradient updates can align with longer term training trajectories.

\section{Acknowledgements}
This material is based upon work supported by the National Science Foundation under Grants No. 2218063 and 2239691. Computations were performed on the Vermont Advanced Computing Core supported in part by NSF Award No. OAC-1827314 and by hardware donations from AMD as part of their HPC Fund.

\bibliography{references}

\begin{thebibliography}{42}
\providecommand{\natexlab}[1]{#1}
\providecommand{\url}[1]{\texttt{#1}}
\expandafter\ifx\csname urlstyle\endcsname\relax
  \providecommand{\doi}[1]{doi: #1}\else
  \providecommand{\doi}{doi: \begingroup \urlstyle{rm}\Url}\fi

\bibitem[Alain \& Bengio(2018)Alain and Bengio]{alain2018understandingintermediatelayersusing}
Guillaume Alain and Yoshua Bengio.
\newblock Understanding intermediate layers using linear classifier probes, 2018.
\newblock URL \url{https://arxiv.org/abs/1610.01644v4}.

\bibitem[Amari \& Douglas(1998)Amari and Douglas]{amari1998natural}
Shun-Ichi Amari and Scott~C Douglas.
\newblock Why natural gradient?
\newblock In \emph{Proceedings of the 1998 IEEE International Conference on Acoustics, Speech and Signal Processing, ICASSP'98 (Cat. No. 98CH36181)}, volume~2, pp.\  1213--1216. IEEE, 1998.

\bibitem[Beaulieu et~al.(2020)Beaulieu, Frati, Miconi, Lehman, Stanley, Clune, and Cheney]{beaulieu2020learningcontinuallylearn}
Shawn Beaulieu, Lapo Frati, Thomas Miconi, Joel Lehman, Kenneth~O. Stanley, Jeff Clune, and Nick Cheney.
\newblock Learning to continually learn, 2020.
\newblock URL \url{https://arxiv.org/abs/2002.09571v2}.

\bibitem[Chen et~al.(2023)Chen, Marchisio, Raileanu, Adelani, Saito~Stenetorp, Riedel, and Artetxe]{chen2023improving}
Yihong Chen, Kelly Marchisio, Roberta Raileanu, David Adelani, Pontus Lars~Erik Saito~Stenetorp, Sebastian Riedel, and Mikel Artetxe.
\newblock Improving language plasticity via pretraining with active forgetting.
\newblock \emph{Advances in Neural Information Processing Systems}, 36:\penalty0 31543--31557, 2023.

\bibitem[Dohare et~al.(2024)Dohare, Hernandez-Garcia, Lan, Rahman, Mahmood, and Sutton]{dohare2024loss}
Shibhansh Dohare, J~Fernando Hernandez-Garcia, Qingfeng Lan, Parash Rahman, A~Rupam Mahmood, and Richard~S Sutton.
\newblock Loss of plasticity in deep continual learning.
\newblock \emph{Nature}, 632\penalty0 (8026):\penalty0 768--774, 2024.

\bibitem[Draxler et~al.(2018)Draxler, Veschgini, Salmhofer, and Hamprecht]{draxler2018essentially}
Felix Draxler, Kambis Veschgini, Manfred Salmhofer, and Fred Hamprecht.
\newblock Essentially no barriers in neural network energy landscape.
\newblock In \emph{International conference on machine learning}, pp.\  1309--1318. PMLR, 2018.

\bibitem[Finn et~al.(2017)Finn, Abbeel, and Levine]{finn2017modelagnosticmetalearning}
Chelsea Finn, Pieter Abbeel, and Sergey Levine.
\newblock Model-agnostic meta-learning for fast adaptation of deep networks.
\newblock In \emph{International conference on machine learning}, pp.\  1126--1135. PMLR, 2017.

\bibitem[Fort et~al.(2020{\natexlab{a}})Fort, Dziugaite, Paul, Kharaghani, Roy, and Ganguli]{fort2020deeplearningversuskernel}
Stanislav Fort, Gintare~Karolina Dziugaite, Mansheej Paul, Sepideh Kharaghani, Daniel~M. Roy, and Surya Ganguli.
\newblock Deep learning versus kernel learning: an empirical study of loss landscape geometry and the time evolution of the neural tangent kernel, 2020{\natexlab{a}}.
\newblock URL \url{https://arxiv.org/abs/2010.15110v1}.

\bibitem[Fort et~al.(2020{\natexlab{b}})Fort, Hu, and Lakshminarayanan]{fort2020deepensembleslosslandscape}
Stanislav Fort, Huiyi Hu, and Balaji Lakshminarayanan.
\newblock Deep ensembles: A loss landscape perspective, 2020{\natexlab{b}}.
\newblock URL \url{https://arxiv.org/abs/1912.02757v2}.

\bibitem[Frankle et~al.(2020)Frankle, Schwab, and Morcos]{frankle2020earlyphaseneuralnetwork}
Jonathan Frankle, David~J. Schwab, and Ari~S. Morcos.
\newblock The early phase of neural network training, 2020.
\newblock URL \url{https://arxiv.org/abs/2002.10365v1}.

\bibitem[Frati et~al.(2023)Frati, Traft, and Cheney]{frati2023omnimage}
Lapo Frati, Neil Traft, and Nick Cheney.
\newblock Omnimage: Evolving 1k image cliques for few-shot learning.
\newblock In \emph{Proceedings of the Genetic and Evolutionary Computation Conference}, pp.\  476--484, 2023.

\bibitem[Frati et~al.(2024)Frati, Traft, Clune, and Cheney]{frati2024resetecai}
Lapo Frati, Neil Traft, Jeff Clune, and Nick Cheney.
\newblock Reset it and forget it: Relearning last-layer weights improves continual and transfer learning.
\newblock In \emph{ECAI 2024}, pp.\  2998--3005. IOS Press, 2024.

\bibitem[Garipov et~al.(2018)Garipov, Izmailov, Podoprikhin, Vetrov, and Wilson]{garipov2018losssurfacesmodeconnectivity}
Timur Garipov, Pavel Izmailov, Dmitrii Podoprikhin, Dmitry Vetrov, and Andrew~Gordon Wilson.
\newblock Loss surfaces, mode connectivity, and fast ensembling of dnns, 2018.
\newblock URL \url{https://arxiv.org/abs/1802.10026v4}.

\bibitem[Hubara et~al.(2018)Hubara, Courbariaux, Soudry, El-Yaniv, and Bengio]{hubara2018quantized}
Itay Hubara, Matthieu Courbariaux, Daniel Soudry, Ran El-Yaniv, and Yoshua Bengio.
\newblock Quantized neural networks: Training neural networks with low precision weights and activations.
\newblock \emph{Journal of Machine Learning Research}, 18\penalty0 (187):\penalty0 1--30, 2018.

\bibitem[Hwang(2024)]{hwang2024fadamadamnaturalgradient}
Dongseong Hwang.
\newblock Fadam: Adam is a natural gradient optimizer using diagonal empirical fisher information, 2024.
\newblock URL \url{https://arxiv.org/abs/2405.12807v11}.

\bibitem[Javed \& White(2019)Javed and White]{javed2019metalearningrepresentationscontinuallearning}
Khurram Javed and Martha White.
\newblock Meta-learning representations for continual learning, 2019.
\newblock URL \url{https://arxiv.org/abs/1905.12588v2}.

\bibitem[Jin et~al.(2020)Jin, Yi, Zhang, Zhang, Schewe, and Huang]{jin2020does}
Gaojie Jin, Xinping Yi, Liang Zhang, Lijun Zhang, Sven Schewe, and Xiaowei Huang.
\newblock How does weight correlation affect generalisation ability of deep neural networks?
\newblock \emph{Advances in Neural Information Processing Systems}, 33:\penalty0 21346--21356, 2020.

\bibitem[Kingma \& Ba(2017)Kingma and Ba]{kingma2017adammethodstochasticoptimization}
Diederik~P. Kingma and Jimmy Ba.
\newblock Adam: A method for stochastic optimization, 2017.
\newblock URL \url{https://arxiv.org/abs/1412.6980v9}.

\bibitem[Kirkpatrick et~al.(2017)Kirkpatrick, Pascanu, Rabinowitz, Veness, Desjardins, Rusu, Milan, Quan, Ramalho, Grabska-Barwinska, et~al.]{kirkpatrick2017overcoming}
James Kirkpatrick, Razvan Pascanu, Neil Rabinowitz, Joel Veness, Guillaume Desjardins, Andrei~A Rusu, Kieran Milan, John Quan, Tiago Ramalho, Agnieszka Grabska-Barwinska, et~al.
\newblock Overcoming catastrophic forgetting in neural networks.
\newblock \emph{Proceedings of the national academy of sciences}, 114\penalty0 (13):\penalty0 3521--3526, 2017.

\bibitem[Krishnamoorthi(2018)]{krishnamoorthi2018quantizing}
Raghuraman Krishnamoorthi.
\newblock Quantizing deep convolutional networks for efficient inference: A whitepaper, 2018.
\newblock URL \url{https://arxiv.org/abs/1806.08342v6}.

\bibitem[Kumar et~al.(2022)Kumar, Raghunathan, Jones, Ma, and Liang]{kumar2022finetuningdistortpretrainedfeatures}
Ananya Kumar, Aditi Raghunathan, Robbie Jones, Tengyu Ma, and Percy Liang.
\newblock Fine-tuning can distort pretrained features and underperform out-of-distribution, 2022.
\newblock URL \url{https://arxiv.org/abs/2202.10054v1}.

\bibitem[Kumaran et~al.(2016)Kumaran, Hassabis, and McClelland]{kumaran2016learning}
Dharshan Kumaran, Demis Hassabis, and James~L McClelland.
\newblock What learning systems do intelligent agents need? complementary learning systems theory updated.
\newblock \emph{Trends in cognitive sciences}, 20\penalty0 (7):\penalty0 512--534, 2016.

\bibitem[Kunstner et~al.(2019)Kunstner, Hennig, and Balles]{kunstner2019limitations}
Frederik Kunstner, Philipp Hennig, and Lukas Balles.
\newblock Limitations of the empirical fisher approximation for natural gradient descent.
\newblock \emph{Advances in neural information processing systems}, 32, 2019.

\bibitem[Lake et~al.(2015)Lake, Salakhutdinov, and Tenenbaum]{lake2015human}
Brenden~M Lake, Ruslan Salakhutdinov, and Joshua~B Tenenbaum.
\newblock Human-level concept learning through probabilistic program induction.
\newblock \emph{Science}, 350\penalty0 (6266):\penalty0 1332--1338, 2015.

\bibitem[Lee et~al.(2020)Lee, Xin, Lee, and Parameswaran]{lee2020demystifying}
Angela Lee, Doris Xin, Doris Lee, and Aditya Parameswaran.
\newblock Demystifying a dark art: Understanding real-world machine learning model development, 2020.
\newblock URL \url{https://arxiv.org/abs/2005.01520v1}.

\bibitem[Li et~al.(2018{\natexlab{a}})Li, Farkhoor, Liu, and Yosinski]{li2018measuringintrinsicdimensionobjective}
Chunyuan Li, Heerad Farkhoor, Rosanne Liu, and Jason Yosinski.
\newblock Measuring the intrinsic dimension of objective landscapes, 2018{\natexlab{a}}.
\newblock URL \url{https://arxiv.org/abs/1804.08838v1}.

\bibitem[Li et~al.(2018{\natexlab{b}})Li, Xu, Taylor, Studer, and Goldstein]{li2018visualizing}
Hao Li, Zheng Xu, Gavin Taylor, Christoph Studer, and Tom Goldstein.
\newblock Visualizing the loss landscape of neural nets.
\newblock \emph{Advances in neural information processing systems}, 31, 2018{\natexlab{b}}.

\bibitem[Lyle et~al.(2023)Lyle, Zheng, Nikishin, Pires, Pascanu, and Dabney]{lyle2023understanding}
Clare Lyle, Zeyu Zheng, Evgenii Nikishin, Bernardo~Avila Pires, Razvan Pascanu, and Will Dabney.
\newblock Understanding plasticity in neural networks.
\newblock In \emph{International Conference on Machine Learning}, pp.\  23190--23211. PMLR, 2023.

\bibitem[Mallya \& Lazebnik(2018)Mallya and Lazebnik]{mallya2018packnet}
Arun Mallya and Svetlana Lazebnik.
\newblock Packnet: Adding multiple tasks to a single network by iterative pruning.
\newblock In \emph{Proceedings of the IEEE conference on Computer Vision and Pattern Recognition}, pp.\  7765--7773, 2018.

\bibitem[Martens(2020)]{martens2020new}
James Martens.
\newblock New insights and perspectives on the natural gradient method.
\newblock \emph{Journal of Machine Learning Research}, 21\penalty0 (146):\penalty0 1--76, 2020.

\bibitem[McClelland et~al.(1995)McClelland, McNaughton, and O'Reilly]{mcclelland1995there}
James~L McClelland, Bruce~L McNaughton, and Randall~C O'Reilly.
\newblock Why there are complementary learning systems in the hippocampus and neocortex: insights from the successes and failures of connectionist models of learning and memory.
\newblock \emph{Psychological review}, 102\penalty0 (3):\penalty0 419, 1995.

\bibitem[Mermillod et~al.(2013)Mermillod, Bugaiska, and Bonin]{mermillod2013stability}
Martial Mermillod, Aur{\'e}lia Bugaiska, and Patrick Bonin.
\newblock The stability-plasticity dilemma: Investigating the continuum from catastrophic forgetting to age-limited learning effects, 2013.

\bibitem[Miconi et~al.(2018)Miconi, Stanley, and Clune]{miconi2018differentiable}
Thomas Miconi, Kenneth Stanley, and Jeff Clune.
\newblock Differentiable plasticity: training plastic neural networks with backpropagation.
\newblock In \emph{International Conference on Machine Learning}, pp.\  3559--3568. PMLR, 2018.

\bibitem[Mirzadeh et~al.(2020)Mirzadeh, Farajtabar, Pascanu, and Ghasemzadeh]{mirzadeh2020understanding}
Seyed~Iman Mirzadeh, Mehrdad Farajtabar, Razvan Pascanu, and Hassan Ghasemzadeh.
\newblock Understanding the role of training regimes in continual learning.
\newblock \emph{Advances in Neural Information Processing Systems}, 33:\penalty0 7308--7320, 2020.

\bibitem[Najarro \& Risi(2020)Najarro and Risi]{najarro2020meta}
Elias Najarro and Sebastian Risi.
\newblock Meta-learning through hebbian plasticity in random networks.
\newblock \emph{Advances in Neural Information Processing Systems}, 33:\penalty0 20719--20731, 2020.

\bibitem[Nikishin et~al.(2022)Nikishin, Schwarzer, D’Oro, Bacon, and Courville]{nikishin2022primacy}
Evgenii Nikishin, Max Schwarzer, Pierluca D’Oro, Pierre-Luc Bacon, and Aaron Courville.
\newblock The primacy bias in deep reinforcement learning.
\newblock In \emph{International conference on machine learning}, pp.\  16828--16847. PMLR, 2022.

\bibitem[Polyak(1964)]{polyak1964some}
Boris~T Polyak.
\newblock Some methods of speeding up the convergence of iteration methods.
\newblock \emph{Ussr computational mathematics and mathematical physics}, 4\penalty0 (5):\penalty0 1--17, 1964.

\bibitem[Rebuffi et~al.(2017)Rebuffi, Kolesnikov, Sperl, and Lampert]{rebuffi2017icarl}
Sylvestre-Alvise Rebuffi, Alexander Kolesnikov, Georg Sperl, and Christoph~H Lampert.
\newblock icarl: Incremental classifier and representation learning.
\newblock In \emph{Proceedings of the IEEE conference on Computer Vision and Pattern Recognition}, pp.\  2001--2010, 2017.

\bibitem[Ulyanov et~al.(2017)Ulyanov, Vedaldi, and Lempitsky]{ulyanov2017instancenormalizationmissingingredient}
Dmitry Ulyanov, Andrea Vedaldi, and Victor Lempitsky.
\newblock Instance normalization: The missing ingredient for fast stylization, 2017.
\newblock URL \url{https://arxiv.org/abs/1607.08022v3}.

\bibitem[Wang et~al.(2024)Wang, Zhang, Su, and Zhu]{continuallearningsurvey}
Liyuan Wang, Xingxing Zhang, Hang Su, and Jun Zhu.
\newblock A comprehensive survey of continual learning: Theory, method and application.
\newblock \emph{IEEE Transactions on Pattern Analysis and Machine Intelligence}, 46\penalty0 (8):\penalty0 5362--5383, 2024.
\newblock \doi{10.1109/TPAMI.2024.3367329}.

\bibitem[Yosinski et~al.(2014)Yosinski, Clune, Bengio, and Lipson]{yosinski2014transferable}
Jason Yosinski, Jeff Clune, Yoshua Bengio, and Hod Lipson.
\newblock How transferable are features in deep neural networks?
\newblock \emph{Advances in neural information processing systems}, 27, 2014.

\bibitem[Zhuang et~al.(2020)Zhuang, Qi, Duan, Xi, Zhu, Zhu, Xiong, and He]{zhuang2020comprehensive}
Fuzhen Zhuang, Zhiyuan Qi, Keyu Duan, Dongbo Xi, Yongchun Zhu, Hengshu Zhu, Hui Xiong, and Qing He.
\newblock A comprehensive survey on transfer learning.
\newblock \emph{Proceedings of the IEEE}, 109\penalty0 (1):\penalty0 43--76, 2020.

\end{thebibliography}
\bibliographystyle{collas2025_conference}

\clearpage 
\appendix
\section{Network Structure}
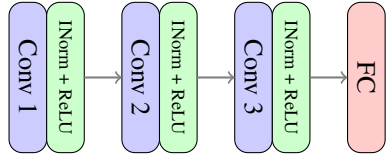
\begin{figure}[!ht]
    \centering
\begin{tikzpicture}
    \draw[fill=blue!20, rounded corners=5pt] (0,0) rectangle (0.5,2);
    \node[rotate=-90] at (0.25, 1) {Conv 1};
    \draw[fill=green!20, rounded corners=5pt] (0.5,0) rectangle (1,2);
    \node[rotate=-90, scale=0.7] at (0.75, 1) {INorm + ReLU};

    \draw[fill=blue!20, rounded corners=5pt] (1.5,0) rectangle (2,2);
    \node[rotate=-90] at (1.75, 1) {Conv 2};
    \draw[fill=green!20, rounded corners=5pt] (2,0) rectangle (2.5,2);
    \node[rotate=-90,, scale=0.7] at (2.25, 1) {INorm + ReLU};

    \draw[fill=blue!20, rounded corners=5pt] (3,0) rectangle (3.5,2);
    \node[rotate=-90] at (3.25, 1) {Conv 3};
    \draw[fill=green!20, rounded corners=5pt] (3.5,0) rectangle (4,2);
    \node[rotate=-90, scale=0.7] at (3.75, 1) {INorm + ReLU};

    \draw[fill=red!20, rounded corners=5pt] (4.5,0) rectangle (5,2);
    \node[rotate=-90] at (4.75, 1) {FC};

    \draw[->, thick, color=gray] (1,1) -- (1.5,1);
    \draw[->, thick, color=gray] (2.5,1) -- (3,1);
    \draw[->, thick, color=gray] (4,1) -- (4.5,1);
\end{tikzpicture}
    \caption{Convnet structure with normalization layers}
    \label{fig:convnet_struct}
\end{figure}

Following \cite{frati2024resetecai} we employ a compact convolutional neural network consisting of three blocks, each containing convolution, InstanceNorm \cite{ulyanov2017instancenormalizationmissingingredient}, ReLU activation, and max pooling layers (except for the final block, which omits pooling). All convolutional layers maintain 256 output channels. The network processes 28×28 single-channel images for Omniglot and 84×84 RGB images for omni-image, with the architecture adjusted accordingly. A single fully-connected layer serves as the classifier (see Figure \ref{fig:convnet_struct}).
Training with the SGD uses momentum $\mu = 0.9$.
Training with the Adam optimizer uses PyTorch's default parameters for betas ($\beta_1 = 0.9, \beta_2 = 0.999)$.

\section{Optimizers \label{appendix:optim}}
\add{Optimizers used are from PyTorch 2.2.0,} SGD with momentum computes a parameter update as 
\add{\begin{align}
    b_t &\leftarrow \mu b_{t-1} + \nabla_\theta f(\theta_{t-1})\\
    \theta_t &\leftarrow \theta_{t - 1} - \gamma \cdot b_t
\end{align}}
where $\gamma$ is the learning rate, and $\mu$ controls the amount of momentum\footnote{The implementation of momentum that PyTorch uses is a modified version of the \cite{polyak1964some} update, see \url{https://pytorch.org/docs/2.2/generated/torch.optim.SGD.html\#sgd} for more information.}.

Adam \cite{kingma2017adammethodstochasticoptimization} computes individual adaptive learning rates for different parameters from estimates of first and second moments of the gradients as follows
\begin{align}
    g_t &\leftarrow \nabla_\theta f(\theta_{t-1})\\
    m_t &\leftarrow  \text{lerp}(\;g_t,\;   m_{t-1},\; \beta_1\;)\\
    v_t &\leftarrow  \text{lerp}(\;g_t^2,\; v_{t-1},\; \beta_2\;)\\
    \theta_t &\leftarrow \theta_{t - 1} + \gamma \cdot \widehat{m_t}/(\sqrt{\widehat{v_t}} + \epsilon) \label{eq:adam:sqrt}
\end{align}
where $\widehat{m_t}$ and $\widehat{v_t}$ are, respectively, the bias-corrected first moment and second raw moments estimates, and $\text{lerp}(a,b,t) = (1-t)\cdot a + t \cdot b$.

\clearpage
\section{Sequential training on Omniglot: All accuracies}

\begin{table*}[!htbp]
\centering
\rowcolors{2}{gray!20}{white}%
\begin{tabular}{llc|cccc}
\toprule
optimizer & zapped & epoch & lr=0.0001 & lr=0.0003 & lr=0.0006 & lr=0.0010 \\
\midrule
\multirow[t]{6}{*}{adam} & \multirow[t]{3}{*}{unzapped} & 0 & 0.197 {\scriptsize $\pm 0.009$} & 0.351 {\scriptsize $\pm 0.011$} & 0.518 {\scriptsize $\pm 0.013$} & 0.664 {\scriptsize $\pm 0.010$} \\
 &  & 1 & 0.458 {\scriptsize $\pm 0.021$} & 0.694 {\scriptsize $\pm 0.014$} & 0.708 {\scriptsize $\pm 0.014$} & 0.708 {\scriptsize $\pm 0.014$} \\
 &  & 2 & 0.636 {\scriptsize $\pm 0.019$} & 0.734 {\scriptsize $\pm 0.010$} & 0.737 {\scriptsize $\pm 0.007$} & 0.743 {\scriptsize $\pm 0.011$} \\
 & \multirow[t]{3}{*}{zapped} & 0 & 0.249 {\scriptsize $\pm 0.009$} & 0.452 {\scriptsize $\pm 0.006$} & 0.630 {\scriptsize $\pm 0.016$} & \textbf{0.802} {\scriptsize $\pm 0.013$} \\
 &  & 1 & 0.591 {\scriptsize $\pm 0.008$} & 0.833 {\scriptsize $\pm 0.008$} & 0.839 {\scriptsize $\pm 0.008$} & 0.839 {\scriptsize $\pm 0.010$} \\
 &  & 2 & 0.796 {\scriptsize $\pm 0.009$} & 0.858 {\scriptsize $\pm 0.011$} & 0.865 {\scriptsize $\pm 0.012$} & \textbf{0.865} {\scriptsize $\pm 0.011$} \\
\multirow[t]{6}{*}{sgd} & \multirow[t]{3}{*}{unzapped} & 0 & 0.077 {\scriptsize $\pm 0.013$} & 0.392 {\scriptsize $\pm 0.026$} & 0.542 {\scriptsize $\pm 0.026$} & 0.569 {\scriptsize $\pm 0.022$} \\
 &  & 1 & 0.256 {\scriptsize $\pm 0.026$} & 0.555 {\scriptsize $\pm 0.016$} & 0.595 {\scriptsize $\pm 0.022$} & 0.585 {\scriptsize $\pm 0.018$} \\
 &  & 2 & 0.389 {\scriptsize $\pm 0.028$} & 0.597 {\scriptsize $\pm 0.015$} & 0.623 {\scriptsize $\pm 0.015$} & 0.635 {\scriptsize $\pm 0.017$} \\
 & \multirow[t]{3}{*}{zapped} & 0 & 0.119 {\scriptsize $\pm 0.018$} & 0.565 {\scriptsize $\pm 0.014$} & 0.732 {\scriptsize $\pm 0.007$} & \textbf{0.733} {\scriptsize $\pm 0.010$} \\
 &  & 1 & 0.390 {\scriptsize $\pm 0.019$} & 0.744 {\scriptsize $\pm 0.013$} & 0.773 {\scriptsize $\pm 0.013$} & 0.739 {\scriptsize $\pm 0.017$} \\
 &  & 2 & 0.575 {\scriptsize $\pm 0.014$} & 0.785 {\scriptsize $\pm 0.012$} & 0.792 {\scriptsize $\pm 0.009$} & \textbf{0.796} {\scriptsize $\pm 0.011$} \\
\bottomrule
\end{tabular}
\caption{Meta-Test accuracy for configurations in the linear probing setting for sequential learning on Omniglot on 100 tasks.In \textbf{bold} we highlight the best accuracy achieved by each optimizer after the first epoch and last epoch. In the linear probing setting where only the last layer is updated Adam + zapping achieves the highest accuracy both at the end of the first epoch and at the end of the last one.}
\label{tab:frozen_acc_10reps}
\end{table*}

\begin{table*}[!htbp]
\centering
\rowcolors{2}{gray!20}{white}%
\begin{tabular}{llc|cccc}
\toprule
optimizer & zapped & epoch & lr=0.0001 & lr=0.0003 & lr=0.0006 & lr=0.0010 \\
\midrule
\multirow[t]{6}{*}{adam} & \multirow[t]{3}{*}{unzapped} & 0 & 0.194 {\scriptsize $\pm 0.012$} & 0.198 {\scriptsize $\pm 0.009$} & 0.119 {\scriptsize $\pm 0.053$} & 0.047 {\scriptsize $\pm 0.033$} \\
 &  & 1 & 0.577 {\scriptsize $\pm 0.014$} & 0.650 {\scriptsize $\pm 0.015$} & 0.535 {\scriptsize $\pm 0.115$} & 0.374 {\scriptsize $\pm 0.124$} \\
 &  & 2 & 0.723 {\scriptsize $\pm 0.012$} & 0.688 {\scriptsize $\pm 0.020$} & 0.579 {\scriptsize $\pm 0.120$} & 0.460 {\scriptsize $\pm 0.132$} \\
 & \multirow[t]{3}{*}{zapped} & 0 & 0.248 {\scriptsize $\pm 0.011$} & \textbf{0.280} {\scriptsize $\pm 0.010$} & 0.169 {\scriptsize $\pm 0.036$} & 0.070 {\scriptsize $\pm 0.029$} \\
 &  & 1 & 0.742 {\scriptsize $\pm 0.011$} & 0.777 {\scriptsize $\pm 0.020$} & 0.661 {\scriptsize $\pm 0.072$} & 0.417 {\scriptsize $\pm 0.113$} \\
 &  & 2 & \textbf{0.858} {\scriptsize $\pm 0.009$} & 0.814 {\scriptsize $\pm 0.015$} & 0.722 {\scriptsize $\pm 0.072$} & 0.604 {\scriptsize $\pm 0.116$} \\
\multirow[t]{6}{*}{sgd} & \multirow[t]{3}{*}{unzapped} & 0 & 0.075 {\scriptsize $\pm 0.019$} & 0.380 {\scriptsize $\pm 0.025$} & 0.475 {\scriptsize $\pm 0.019$} & 0.480 {\scriptsize $\pm 0.023$} \\
 &  & 1 & 0.258 {\scriptsize $\pm 0.024$} & 0.471 {\scriptsize $\pm 0.026$} & 0.378 {\scriptsize $\pm 0.014$} & 0.418 {\scriptsize $\pm 0.016$} \\
 &  & 2 & 0.401 {\scriptsize $\pm 0.023$} & 0.452 {\scriptsize $\pm 0.015$} & 0.524 {\scriptsize $\pm 0.025$} & 0.580 {\scriptsize $\pm 0.017$} \\
 & \multirow[t]{3}{*}{zapped} & 0 & 0.124 {\scriptsize $\pm 0.018$} & 0.568 {\scriptsize $\pm 0.023$} & 0.639 {\scriptsize $\pm 0.017$} & \textbf{0.655} {\scriptsize $\pm 0.017$} \\
 &  & 1 & 0.400 {\scriptsize $\pm 0.024$} & 0.608 {\scriptsize $\pm 0.019$} & 0.531 {\scriptsize $\pm 0.013$} & 0.599 {\scriptsize $\pm 0.013$} \\
 &  & 2 & 0.582 {\scriptsize $\pm 0.018$} & 0.620 {\scriptsize $\pm 0.011$} & 0.747 {\scriptsize $\pm 0.011$} & \textbf{0.767} {\scriptsize $\pm 0.007$} \\
\bottomrule
\end{tabular}
\caption{Meta-Test accuracy for configurations in the full-model training setting for sequential learning on Omniglot on 100 tasks. In \textbf{bold} we highlight the best accuracy achieved by each optimizer after the first epoch and last epoch. While SGD is significantly better on at the end of the first epoch, Adam achieves better accuracy after 2 more epochs of training, but it requires much lower learning rates. In this low learning rate regime Adam's tracked statistics continue improving previous tasks long after training has ended (see Fig. \ref{fig:sub:unfrozen_adam_low}).}
\label{tab:unfrozen_acc_10reps}
\end{table*}

\clearpage
\section{Zap-Divergence: All Layers \label{appendix:cosim}}

\begin{figure}[!htbp]
\centering
\begin{subfigure}[b]{.45\linewidth}
    \includegraphics[width=\linewidth]{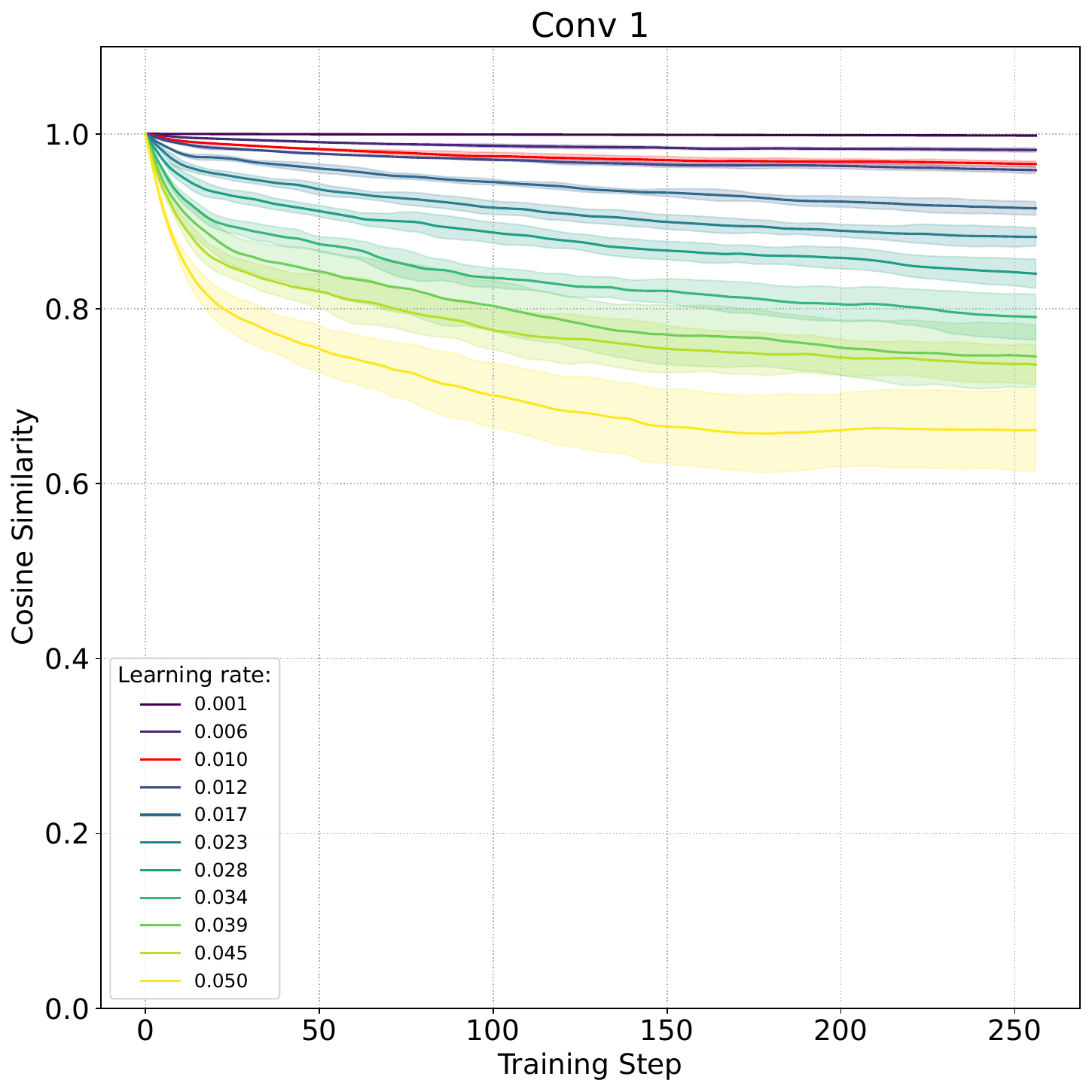}
    \caption{First conv. layer}
    \label{app:sub:cosim_0}
\end{subfigure}
\begin{subfigure}[b]{.45\linewidth}
    \includegraphics[width=\linewidth]{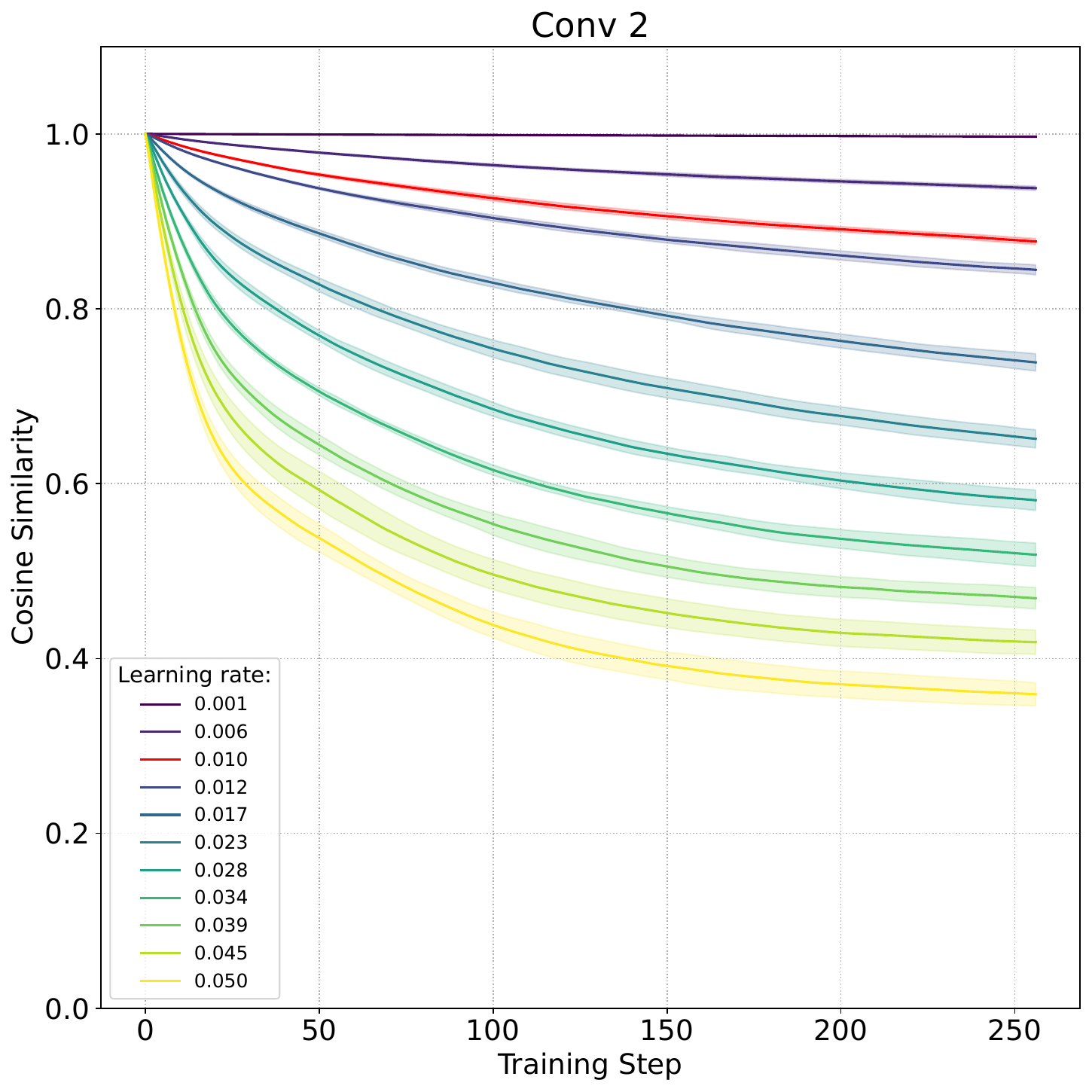}
    \caption{Second conv. layer}
    \label{app:sub:cosim_1}
\end{subfigure}
~ \\
~ \\
~ \\
~ \\
\begin{subfigure}[b]{.45\linewidth}
    \includegraphics[width=\linewidth]{figures/001_A/cosim_lr_encoder.2.pdf}
    \caption{Third conv. layer (same as Fig. \ref{fig:cosim_encoder2})}
    \label{app:sub:cosim_2}
\end{subfigure}
\begin{subfigure}[b]{.45\linewidth}
    \includegraphics[width=\linewidth]{figures/001_A/cosim_lr_fc.pdf}
    \caption{Fully connected layer (same as Fig. \ref{fig:cosim_fc})}
    \label{app:sub:cosim_fc}
\end{subfigure}
\caption{Cosine similarities for all layers, for multiple learning rates.}
\label{app:fig:all_cosims}
\end{figure}

\clearpage
\section{Continual Transfer: Adam\label{appendix:adam}}

\begin{figure}[!htb]
    \centering
    \includegraphics[width=0.7\linewidth]{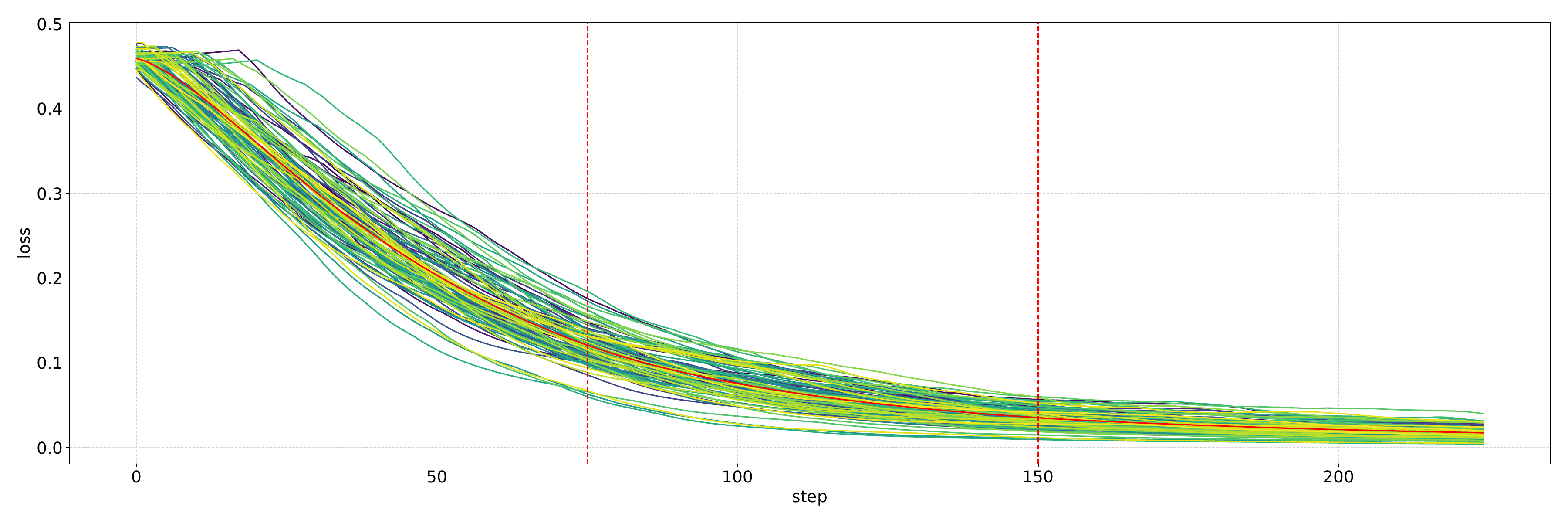}
    \caption{Per-task loss plot in an IID setting (vs sequential in other plots) at transfer time: Since batches contain a random sample of all tasks the loss decreases more uniformly.}
    \label{fig:iid_spaghetti}
\end{figure}

\begin{figure}[!htbp]
\centering
\begin{subfigure}[b]{0.8\linewidth}
    \includegraphics[width=\linewidth]{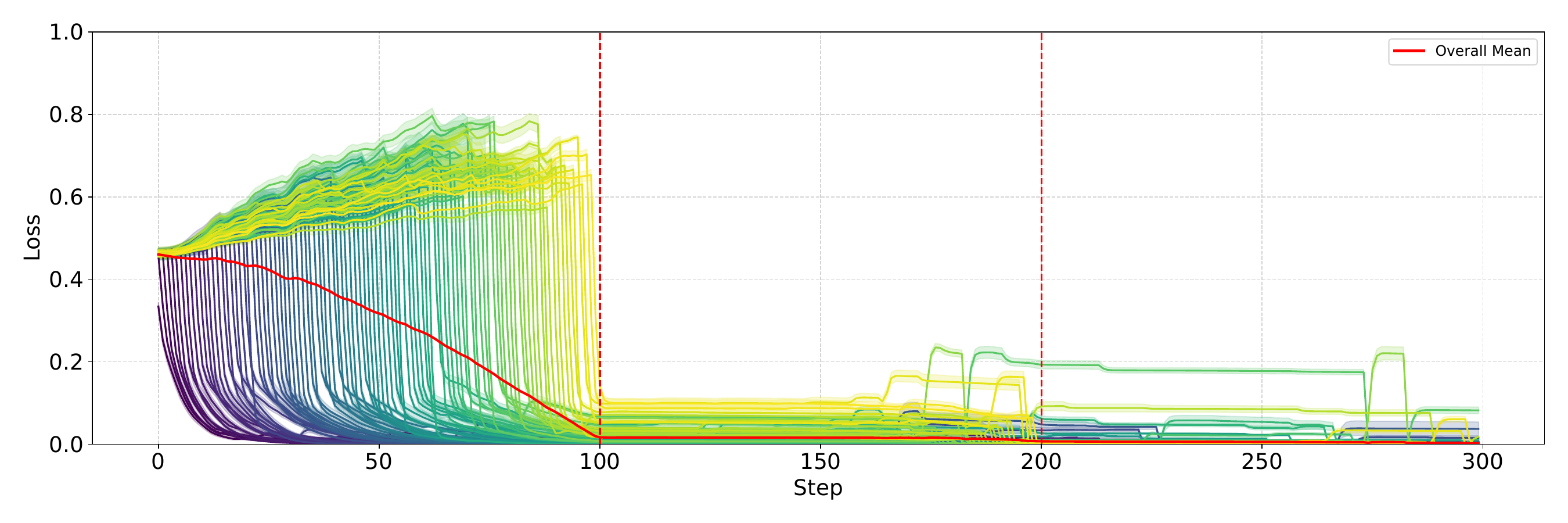}
    \caption{$\gamma = 0.0010$}
    \label{fig:sub:frozen_adam_high}
\end{subfigure}
\begin{subfigure}[b]{0.8\linewidth}
    \includegraphics[width=\linewidth]{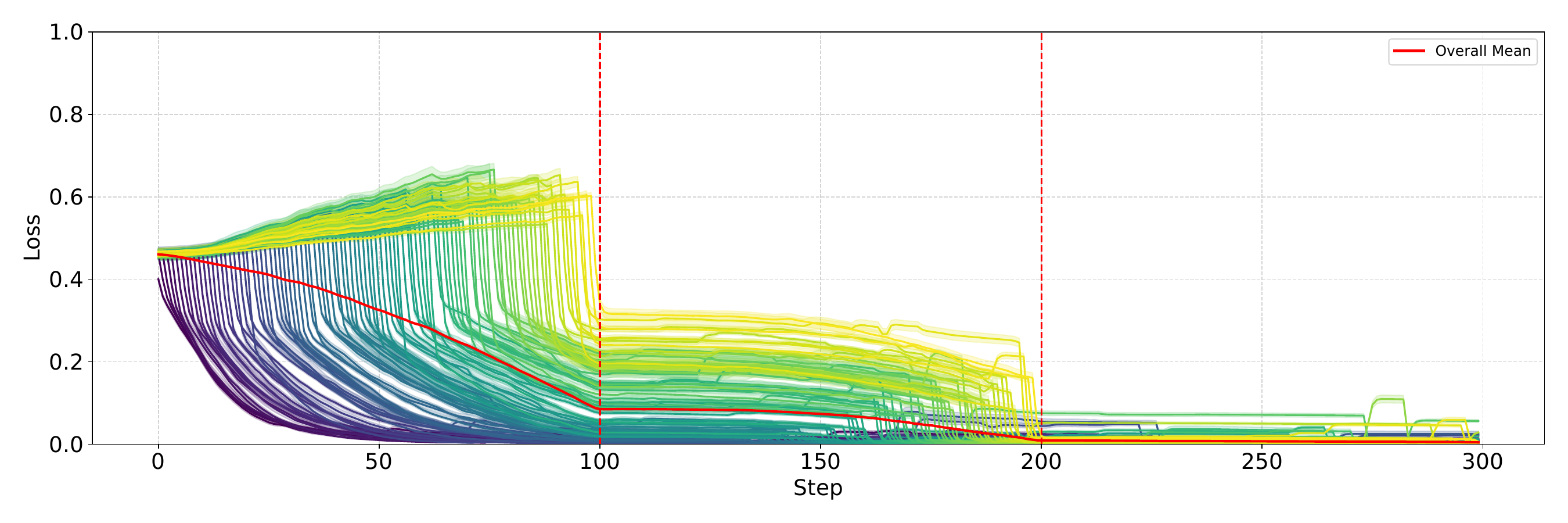}
    \caption{$\gamma = 0.0005$}
    \label{fig:sub:frozen_adam_medium}
\end{subfigure}
\begin{subfigure}[b]{0.8\linewidth}
    \includegraphics[width=\linewidth]{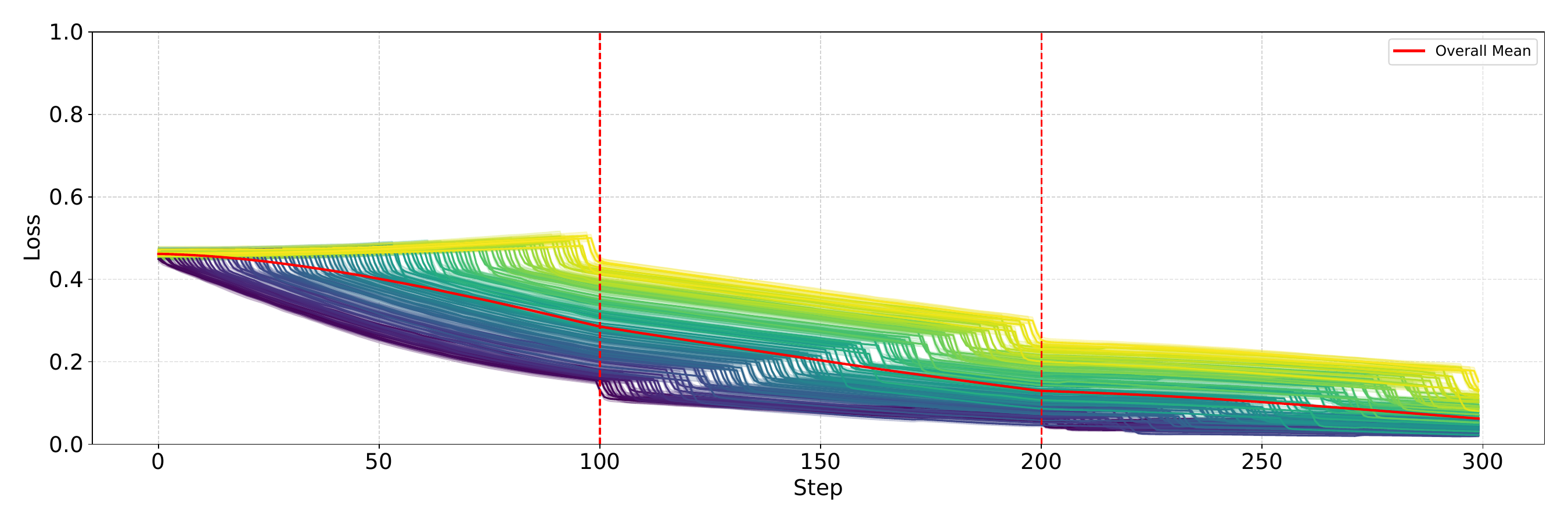}
    \caption{$\gamma = 0.0001$}
    \label{fig:sub:frozen_adam_low}
\end{subfigure}
\caption{Different training dynamics for various learning rates \textit{using linear probing} with Adam on continual learning, with a model that underwent zapping during training.}
\label{fig:frozen_adam_comparison}
\end{figure}

\begin{figure}[!htbp]
\centering
\begin{subfigure}[b]{.8\linewidth}
    \includegraphics[width=\linewidth]{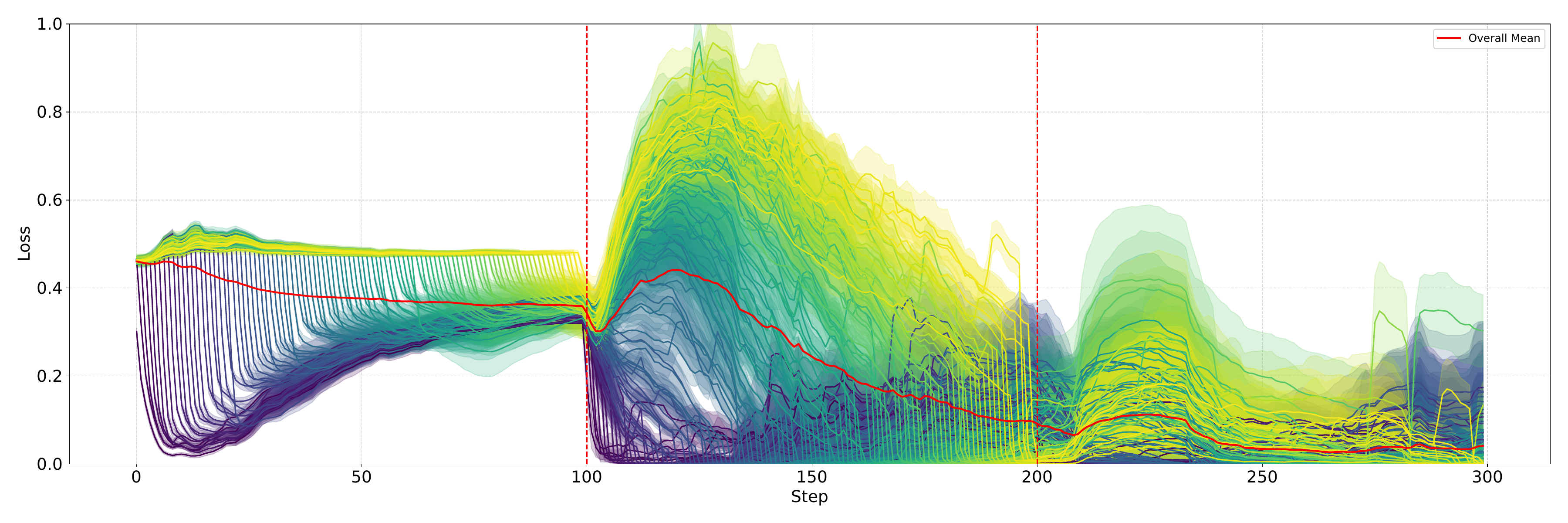}
    \caption{$\gamma = 0.0010$}
    \label{fig:sub:unfrozen_adam_high}
\end{subfigure}
\begin{subfigure}[b]{.8\linewidth}
    \includegraphics[width=\linewidth]{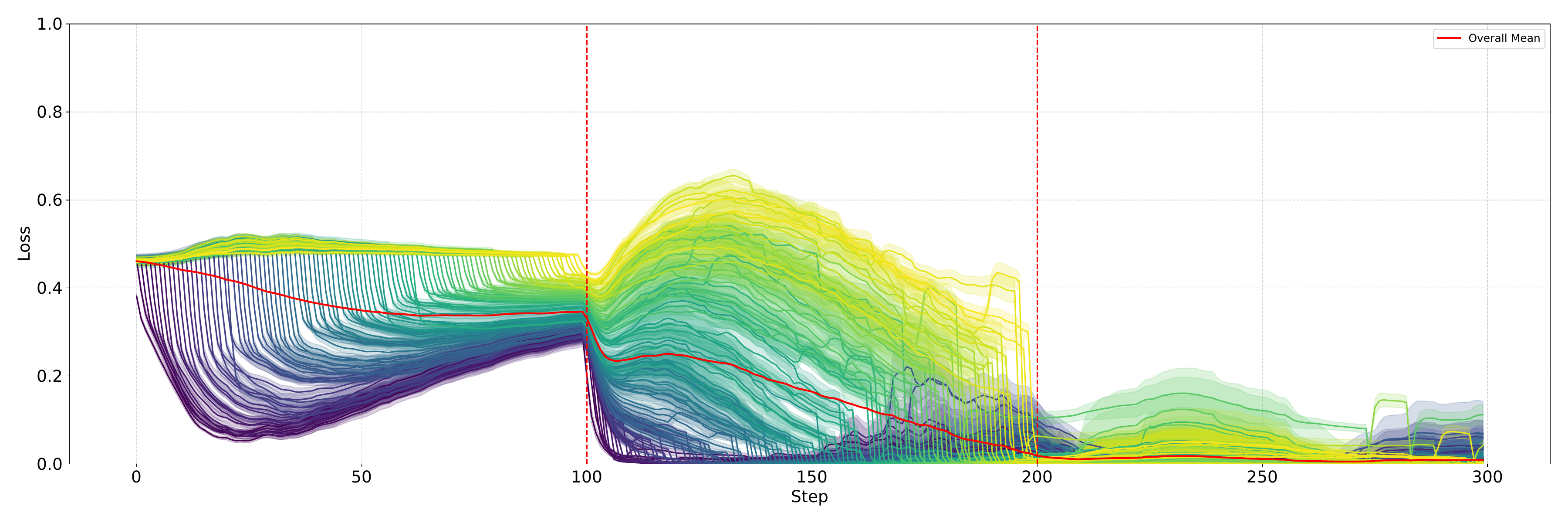}
    \caption{$\gamma = 0.0005$}
    \label{fig:sub:unfrozen_adam_medium}
\end{subfigure}
\begin{subfigure}[b]{.8\linewidth}
    \includegraphics[width=\linewidth]{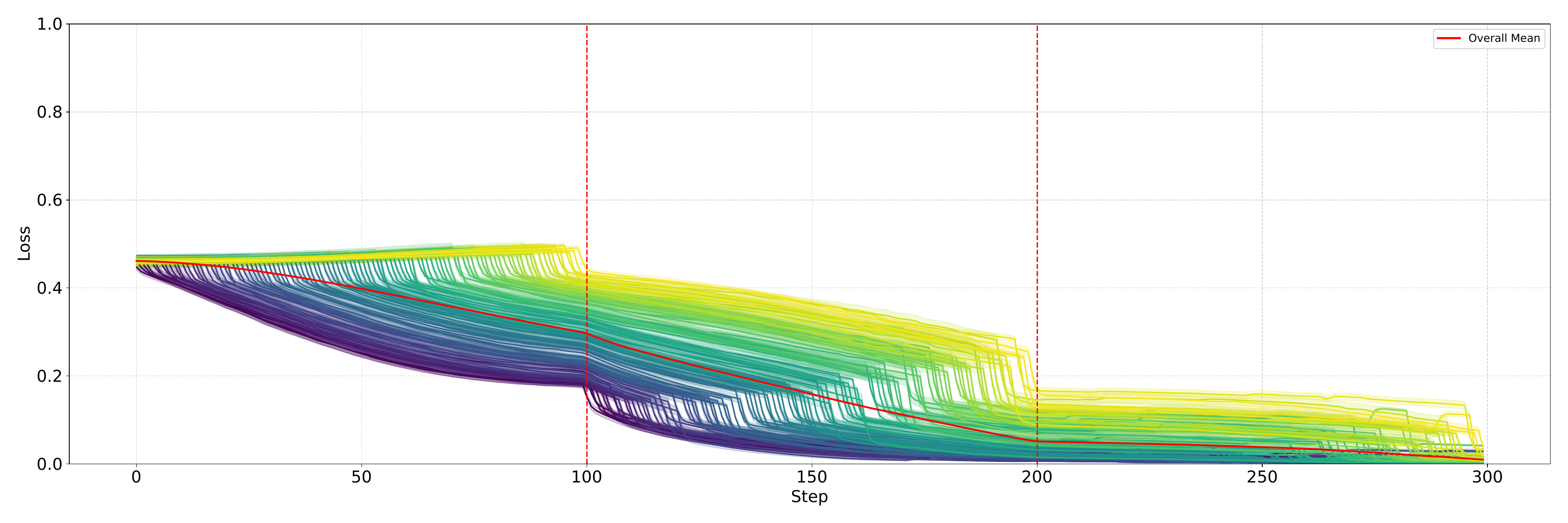}
    \caption{$\gamma = 0.0001$}
    \label{fig:sub:unfrozen_adam_low}
\end{subfigure}
\caption{Different training dynamics for various learning rates \textit{using full-model tuning} with Adam on continual learning, with a model that underwent zapping during training.}
\label{fig:unfrozen_adam_comparison}
\end{figure}

\begin{figure}[!htbp]
\centering
\begin{subfigure}[b]{.9\linewidth}
    \centering
    \includegraphics[trim={0 0 1650pt 0}, clip, height=5cm]{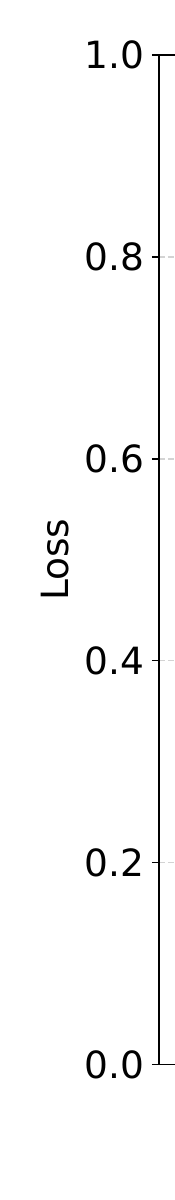}%
    \includegraphics[trim={630pt 0 0 0}, clip, height=5cm]{figures/003_spaghetti/frozen/3epochs/frozen_adam_unzapped_0.0010_3_mean.pdf}
    \caption{Linear probing tuning - unzapped pre-training}
    \label{fig:sub:zap_froz}
\end{subfigure}
\vfill
\begin{subfigure}[b]{.9\linewidth}
    \centering
    \includegraphics[trim={0 0 1220pt 0}, clip, height=5cm]{figures/003_spaghetti/frozen/3epochs/frozen_adam_zapped_0.0010_3_mean.pdf}%
    \includegraphics[trim={480pt 0 0 0}, clip, height=5cm]{figures/003_spaghetti/frozen/3epochs/frozen_adam_zapped_0.0010_3_mean.pdf}
    \caption{Linear probing - zapped pre-training}
    \label{fig:sub:unzap_froz}
\end{subfigure}
\begin{subfigure}[b]{.4\linewidth}
    \centering
    \includegraphics[trim={0 0 1650pt 0}, clip, height=5cm]{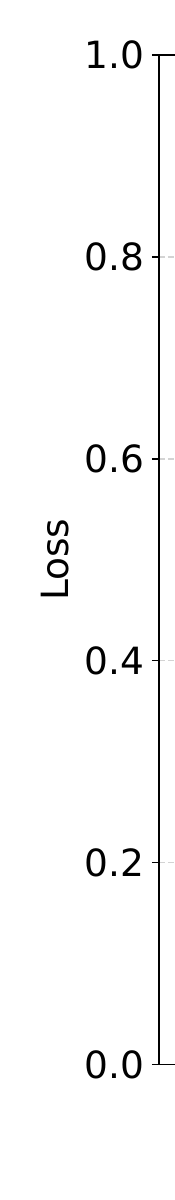}%
    \includegraphics[trim={1130pt 0 0 0}, clip, height=5cm]{figures/003_spaghetti/unfrozen/3epochs/unfrozen_adam_unzapped_0.0010_3_mean.pdf}
    \caption{Full-model tuning - unzapped pre-training}
    \label{fig:sub:zap_unfroz}
\end{subfigure}
\begin{subfigure}[b]{.4\linewidth}
    \centering
    \includegraphics[trim={0 0 1650pt 0}, clip, height=5cm]{figures/003_spaghetti/unfrozen/3epochs/unfrozen_adam_unzapped_0.0010_3_mean.pdf}%
    \includegraphics[trim={1130pt 0 0 0}, clip, height=5cm]{figures/003_spaghetti/unfrozen/3epochs/unfrozen_adam_zapped_0.0010_3_mean.pdf}
    \caption{Full-model tuning - zapped pre-training}
    \label{fig:sub:unzap_unfroz}
\end{subfigure}
\caption{Comparing the effect of zapping on per-task losses: i) in the linear probing case learning is faster and losses more quickly settle on very low values ( Fig. \ref{fig:sub:zap_froz} vs Fig. \ref{fig:sub:unzap_froz} ), ii) in the full-model tuning interference between tasks is reduced, which is particularly noticeable at higher learning rates (Fig. \ref{fig:sub:zap_unfroz} vs \ref{fig:sub:unzap_unfroz}).}
\label{fig:zap_effect_pertask}
\end{figure}

\clearpage
\section{Continual Transfer: SGD\label{appendix:sgd}}

\begin{figure}[!htbp]
\centering
\begin{subfigure}[b]{.8\linewidth}
    \includegraphics[width=\linewidth]{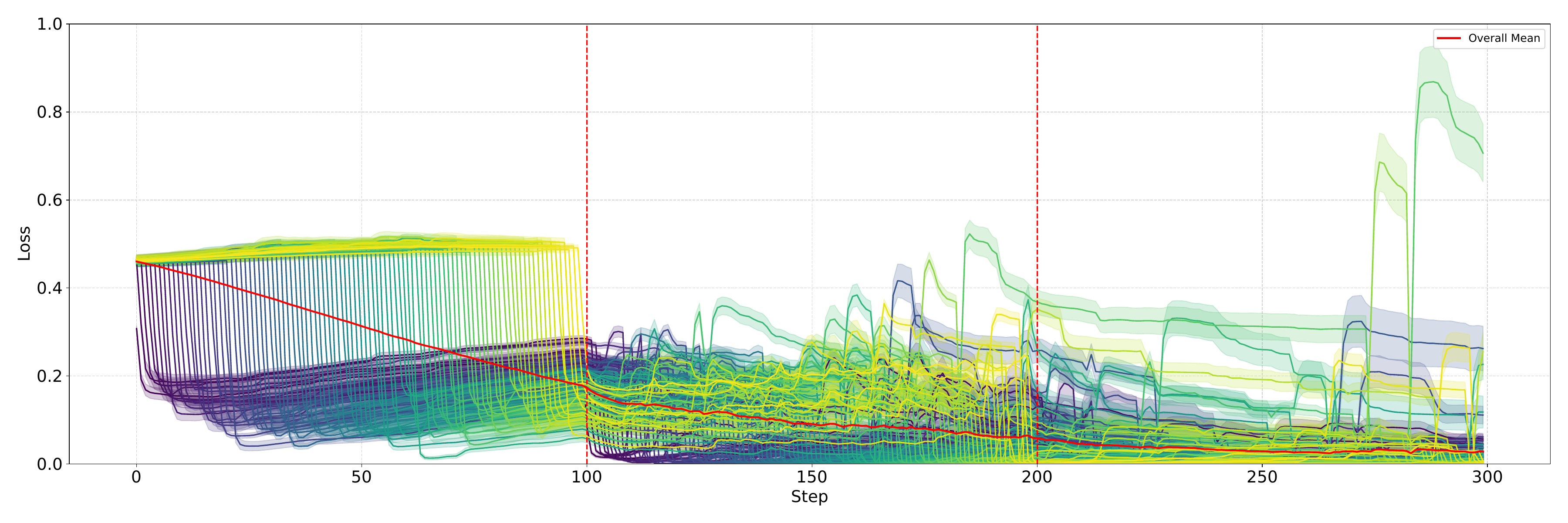}
    \caption{$\gamma = 0.0010$}
    \label{fig:sub:unfrozen_sgd_high}
\end{subfigure}
\begin{subfigure}[b]{.8\linewidth}
    \includegraphics[width=\linewidth]{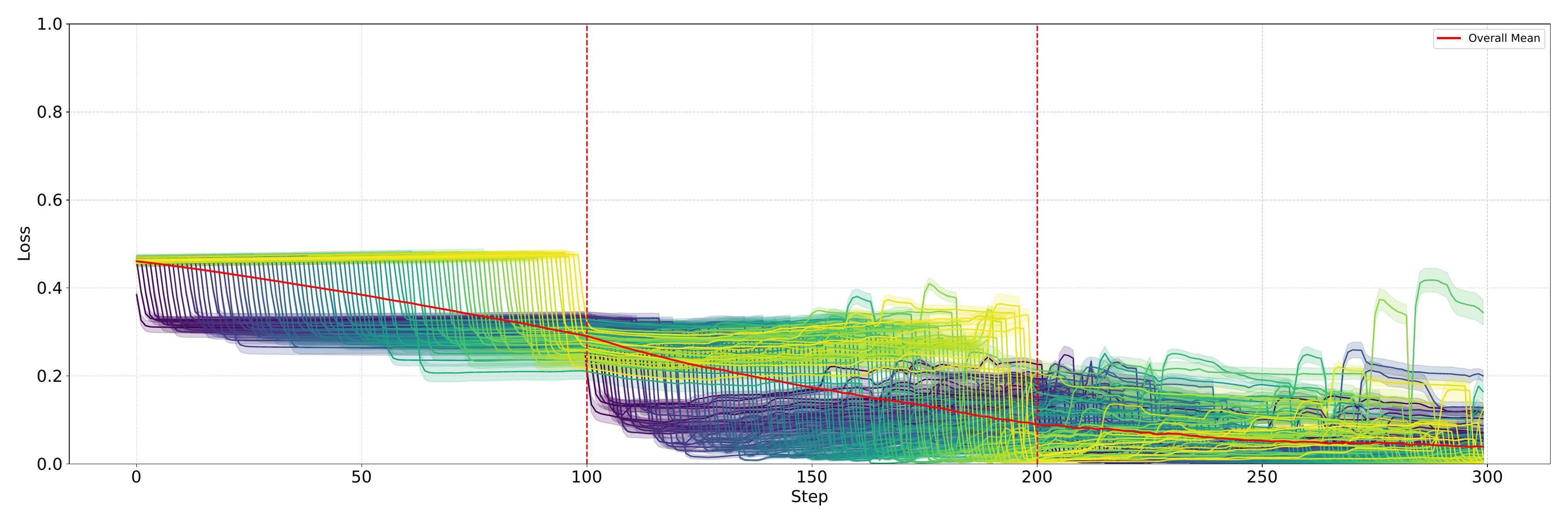}
    \caption{$\gamma = 0.0005$}
    \label{fig:sub:unfrozen_sgd_medium}
\end{subfigure}
\begin{subfigure}[b]{.8\linewidth}
    \includegraphics[width=\linewidth]{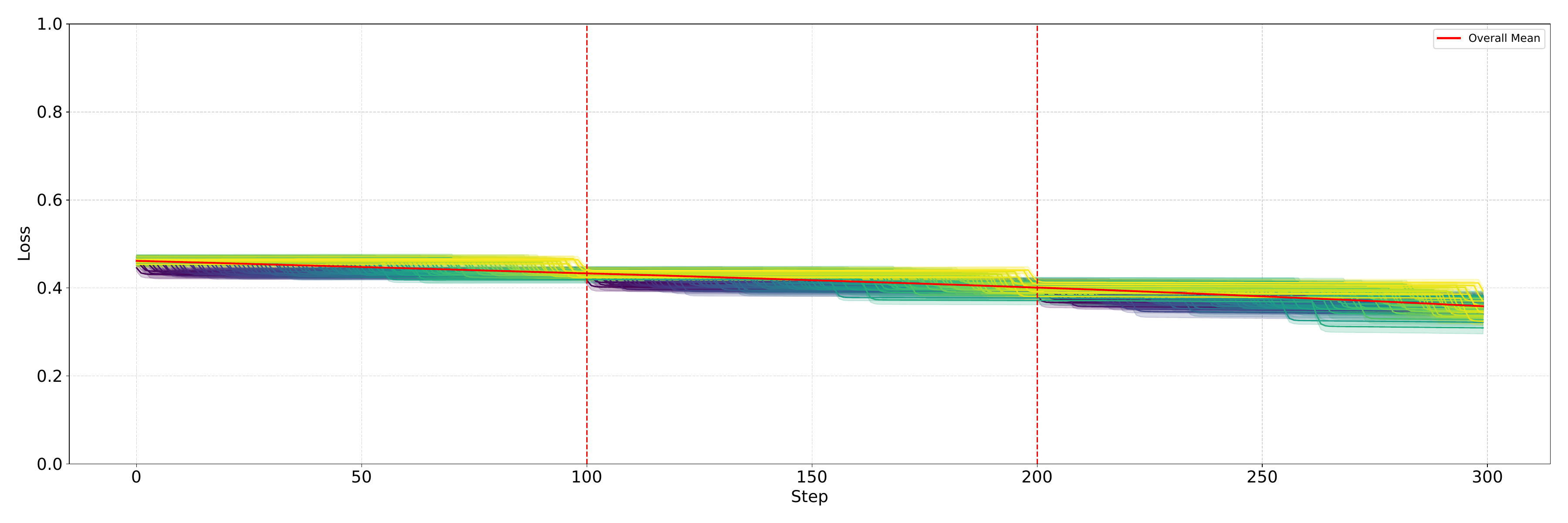}
    \caption{$\gamma = 0.0001$}
    \label{fig:sub:unfrozen_sgd_low}
\end{subfigure}
\caption{Different training dynamics for various learning rates \textit{using full-model tuning} with SGD at Transfer time, on a model that underwent zapping during training.}
\label{fig:unfrozen_sgd_comparison}
\end{figure}

\clearpage
\section{IID Transfer on Omni-image: transfer-train and transfer-test}
\begin{figure}[!htbp]
    \centering
    \includegraphics[width=\linewidth]{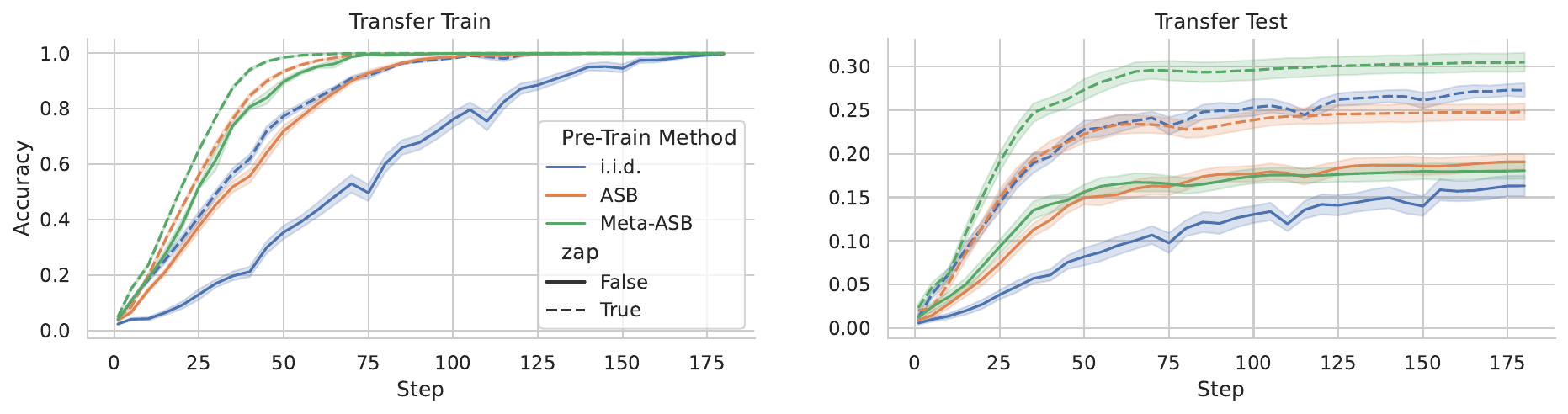}
    \caption{Training (left) and test (right) accuracy on classes during standard fine-tuning on the omni-image subset of ImageNet (15 training images / 5 test images per class). Models pre-trained \textbf{with zapping} achieve significantly higher test accuracies, with models employing both meta-gradients and zapping coming out on top. 
    }
    \label{fig:omni-image-iid_train_and_test}
\end{figure}

\end{document}